\documentclass{sig-alternate}

\usepackage[
pdftitle={Digital Ecosystems: Stability of Evolving Agent Populations},
pdfauthor={Gerard Briscoe},
pdfsubject={Digital Ecosystems},
pdfkeywords={Digital Ecosystems, evolution, stability, agent, entropy, equilibrium},
colorlinks=true, linkcolor=black, citecolor=black, filecolor=black, urlcolor=black, hyperfootnotes=false]{hyperref} 

\usepackage{graphicx, acronym, ifthen, substr, color}
\usepackage[paper=a4paper, asymmetric, left=2cm,right=1.4cm,top=2cm,bottom=4.2cm]{geometry}

\usepackage{amsmath}

\newboolean{final}\setboolean{final}{true}

\newboolean{mactex}\setboolean{mactex}{true}

\newboolean{arxiv}\setboolean{arxiv}{false}

\ifthenelse{\boolean{final}}{}{\usepackage{showlabels}\usepackage{citeext}}

\ifthenelse{\boolean{arxiv}}
	{\newcommand{\arxivfootnote}{\footnote}}
	{\def\arxivfootnote#1{}}

\ifthenelse{\boolean{arxiv}}
	{\def\arxivOnly#1{#1}}
	{\def\arxivOnly#1{}}

\ifthenelse{\boolean{arxiv}}
	{\def\arxivNot#1{}}
	{\def\arxivNot#1{#1}}

\ifthenelse{\boolean{arxiv}}
	{}
	{}

\newcommand{\tfigure}[9]
	{
	\IfSubStringInString{!}{#7}{\begin{figure}[#7]}{\begin{figure}[!t]}
	\IfSubStringInString{mm}{#8}{\vspace{#8}}{}
	\centering

	\IfSubStringInString{pdf}{#3}
		{
		\ifthenelse{\boolean{mactex}}{}{\execute{cd images; ln -s #2.pdf .#2.gdf}}
		\includegraphics[#1]{images/#2}
		}
		{
		\ifthenelse{\boolean{mactex}}{}{\execute{cd images; ./pdfcrop.sh #2}}
		\includegraphics[#1]{images/#2-crop.pdf}
		}

	\vspace{#6}
	\caption[#4]
		{
		\label{#2}
		#4: #5
		}
	\IfSubStringInString{mm}{#9}{\vspace{#9}}{}
	\end{figure}
	}

\newcommand{\Circlesub}[4]
	{
	\ifthenelse{\boolean{mactex}}{}{\immediate\write18{cd images; ./pdfcrop.sh circle#2}}
	\ifthenelse{\boolean{final}}
		{\hspace{#1}\raisebox{#4}{$\includegraphics[scale=0.5, clip=true, trim=0mm 0mm 0mm 0mm]{images/circle#2-crop.pdf}$}\hspace{#3}}
		{\href{file://localhost/Users/g/Desktop/PhDthesis/images/circle#2.graffle}{\hspace{#1}\raisebox{#4}{$\includegraphics[clip=true, trim=0mm 0mm 0mm 0mm]{images/circle#2-crop.pdf}$}\hspace{#3}}}
	}

\newcommand{\execute}[1]{\immediate\write18{#1}}

\definecolor{tred}{RGB}{255,0,0}

\newcommand{\setCap}[2]{#1\immediate\write18{./mkcaption.sh #2}}
\newcommand{\getCap}[1]{\acl*{#1}}

\acrodef{PCG}{Projected Conjugate Gradient} 
\acrodef{QP}{quadratic programming}
\acrodef{RBF}{Radial-Basis Function}
\acrodef{ABM}{Agent-Based Modelling}
\acrodef{AI}{Artificial Intelligence}
\acrodef{DAI}{Distributed Artificial Intelligence}
\acrodef{API}{Application Programming Interface}
\acrodef{ARF}{p14ARF human tumor-suppressor gene}
\acrodef{B2B}{business-to-business}
\acrodef{BDP}{Biological Design Pattern}
\acrodef{BGS}{Best Guess Solution}
\acrodef{BIC}{Biologically-Inspired Computing}
\acrodef{BML}{Business Modelling Language}
\acrodef{BPEL}{Business Process Execution Language}
\acrodef{BPMN}{Business Process Modelling Notation}
\acrodef{CAS}{Complex Adaptive Systems}
\acrodef{COBOL}{COmmon Business-Oriented Language}
\acrodef{DBE}{Digital Business Ecosystem}
\acrodef{DE}{Digital Ecosystem}
\acrodef{DEC}{distributed evolutionary computing}
\acrodef{DGA}{Distributed genetic algorithms}
\acrodef{DIS}{Distributed Intelligence System}
\acrodef{DNA}{Deoxyribose Nucleic Acid}
\acrodef{DOP}{DBE Open Protocol}
\acrodef{DSS}{Distributed Storage System}
\acrodef{EAP}{Evolving Agent Population}
\acrodef{ebXML}{e-business eXtensible Markup Language}
\acrodef{EC}{Evolutionary Computing}
\acrodef{ECJ}{Evolutionary Computing in Java}
\acrodef{EE}{Evolutionary Environment}
\acrodef{EFL}{Evolutionary Framework for Language}
\acrodef{FLE}{Framework for Language Ecosystems}
\acrodef{EOA}{Ecosystem-Oriented Architecture}
\acrodef{ESS}{evolutionary stable strategy}
\acrodef{EvE}{Evolutionary Environment}
\acrodef{ExE}{Execution Environment}
\acrodef{FCB}{Framework for Computational Biomimicry}
\acrodef{FFF}{Fitness Function Framework}
\acrodef{FL}{Fitness Landscape}
\acrodef{HWU}{Heriot-Watt University}
\acrodef{ICL}{Imperial College London}
\acrodef{ICT}{Information and Communications Technology}
\acrodef{INTEL}{Intel Ireland}
\acrodef{IPA}{International Phonetic Alphabet}
\acrodef{ISUFI}{Istituto Superiore Universitario di Formazione Interdisciplinare}
\acrodef{JDJ}{Java Developer's Journal}
\acrodef{KC}{Kolmogorov-Chaitin}
\acrodef{LAN}{local area network}
\acrodef{LSE}{London School of Economics and Political Science}
\acrodef{MAS}{Multi-Agent System}
\acrodef{MDL}{Minimum Description Length}
\acrodef{MDM2}{murine double minute 2}
\acrodef{MFT}{Mean Field Theory}
\acrodef{MoAS}{Mobile Agent System}
\acrodef{MOF}{Meta Object Facility}
\acrodef{MUH}{migration and usage history}
\acrodef{NIC}{Nature Inspired Computing}
\acrodef{NN}{Neural Network}
\acrodef{NoE}{Network of Excellence}
\acrodef{OMG}{Open Mac Grid}
\acrodef{OPAALS}{Open Philosophies for Associative Autopoietic Digital Ecosystems}
\acrodef{P2P}{peer-to-peer}
\acrodef{P53}{protein 53}
\acrodef{PDA}{Personal Digital Assistant}
\acrodef{QoS}{quality of service}
\acrodef{REST}{REpresentational State Transfer}
\acrodef{RNA}{Deoxyribose Nucleic Acid}
\acrodef{SAE}{Software Agent Ecosystem}
\acrodef{SBML}{Systems Biology Modelling Language}
\acrodef{SBVR}{Semantics of Business Vocabulary and Business Rules}
\acrodef{SDL}{Service Description Language}
\acrodef{SF}{Service Factory}
\acrodef{SIM}{Social Interaction Mechanism}
\acrodef{SM}{Service Manifest}
\acrodef{SME}{Small and Medium sized Enterprise}
\acrodef{SML}{Service Modelling Language}
\acrodef{SMO}{Sequential Minimal Optimisation}
\acrodef{SOA}{Service-Oriented Architecture}
\acrodef{SOAP}{Simple Object Access Protocol}
\acrodef{SOC}{Self-Organised Criticality}
\acrodef{SOLUTA}{SOLUTA.NET}
\acrodef{SOM}{Self-Organising Map}
\acrodef{SSL}{Semantic Service Language}
\acrodef{STU}{Salzburg Technical University}
\acrodef{SUN}{Sun Microsystems}
\acrodef{SVM}{Support Vector Machine}
\acrodef{TM}{Turing Machine}
\acrodef{UBHAM}{University of Birmingham}
\acrodef{UDDI}{Universal Description Discovery and Integration}
\acrodef{UML}{Unified Modelling Language}
\acrodef{URI}{Uniform Resource Identifier}
\acrodef{UTM}{Universal Turing Machine}
\acrodef{VLP}{variable length population}
\acrodef{VLS}{variable length sequences}
\acrodef{vls}{variable length sequence}
\acrodef{WP}{Work-Package}
\acrodef{WSDL}{Web Services Definition Language}
\acrodef{XMI}{XML Metadata Interchange}
\acrodef{XML}{eXtensible Markup Language}
\acrodef{MD5}{Message-Digest algorithm 5}
\acrodef{GA}{genetic algorithm}
\acrodef{GP}{genetic programming}
\acrodef{MASON}{Multi-Agent Simulator Of Neighbourhoods}
\acrodef{Repast}{Recursive Porous Agent Simulation Toolkit}
\acrodef{JCLEC}{Java Computing Library for Evolutionary Computing}
\acrodef{OWL-S}{Web Ontology Language - Service}
\acrodef{EGT}{Evolutionary Game Theory}
\acrodef{RBF}{Radial Basis Functions}
\acrodef{SWS}{Semantic Web Services}
\acrodef{HDD}{Hard Disk Drive}
\acrodef{SSD}{Solid-State Drive}
\acrodef{OKS}{Open Knowledge Space}
\acrodef{CAES}{Complex Adaptive EcoSystem}
\acrodef{SaaS}{Software-as-a-Service}
\acrodef{PaaS}{Platform-as-a-Service}
\acrodef{IaaS}{Infrastructure-as-a-Service}
\acrodef{C3}{Community Cloud Computing}

\acrodef{digEco}{with the agents, the populations, the agent migration for \acl{DEC}, and the environmental selection pressures provided by the user base, then the union of the habitats creates the Digital Ecosystem}
\acrodef{archComTop}{many strongly connected clusters (communities), called {sub-networks} (quasi-complete graphs), with a few connections between these clusters (communities). Graphs with this topology have a very high clustering coefficient and small characteristic path lengths}
\acrodef{similarCap}{requests are evaluated on separate {islands} (populations), and so adaptation is accelerated by the sharing of solutions between evolving populations (islands), because they are working to solve similar requests (problems).}
\acrodef{picUser}{will formulate queries to the Digital Ecosystem by creating a request as a {semantic description}, like those being used and developed in \acp{SOA}}
\acrodef{picUserReq}{A population is then instantiated in the user's habitat in response to the user's request, seeded from the agents available at their habitat.}
\acrodef{statesCap}{possible evolutionary path through the state-space $I$}
\acrodef{capStates3}{the {selection pressure} of the evolutionary process}
\acrodef{capStates}{driving it towards the {maximal state} of the {maximum macro-state} $M_{max}$}
\acrodef{graphCap}{in the {maximum macro-state} $M_{max}$ only after generation 178 and always after generation 482.}
\acrodef{aScap}{With the mutation rate under or equal to 60\%, the evolving agent population showed no instability}
\acrodef{digEco}{with the agents, the populations, the agent migration for \acl{DEC}, and the environmental selection pressures provided by the user base, then the union of the habitats creates the Digital Ecosystem}
\acrodef{archComTop}{many strongly connected clusters (communities), called {sub-networks} (quasi-complete graphs), with a few connections between these clusters (communities). Graphs with this topology have a very high clustering coefficient and small characteristic path lengths}
\acrodef{similarCap}{requests are evaluated on separate {islands} (populations), and so adaptation is accelerated by the sharing of solutions between evolving populations (islands), because they are working to solve similar requests (problems).}
\acrodef{picUser}{will formulate queries to the Digital Ecosystem by creating a request as a {semantic description}, like those being used and developed in \acp{SOA}}
\acrodef{picUserReq}{A population is then instantiated in the user's habitat in response to the user's request, seeded from the agents available at their habitat.}
\acrodef{statesCap}{possible evolutionary path through the state-space $I$}
\acrodef{capStates3}{the {selection pressure} of the evolutionary process}
\acrodef{capStates}{driving it towards the {maximal state} of the {maximum macro-state} $M_{max}$}
\acrodef{graphCap}{in the {maximum macro-state} $M_{max}$ only after generation 178 and always after generation 482.}
\acrodef{aScap}{With the mutation rate under or equal to 60\%, the evolving agent population showed no instability}
\acrodef{digEco}{with the agents, the populations, the agent migration for \acl{DEC}, and the environmental selection pressures provided by the user base, then the union of the habitats creates the Digital Ecosystem}
\acrodef{archComTop}{many strongly connected clusters (communities), called {sub-networks} (quasi-complete graphs), with a few connections between these clusters (communities). Graphs with this topology have a very high clustering coefficient and small characteristic path lengths}
\acrodef{similarCap}{requests are evaluated on separate {islands} (populations), and so adaptation is accelerated by the sharing of solutions between evolving populations (islands), because they are working to solve similar requests (problems).}
\acrodef{picUser}{will formulate queries to the Digital Ecosystem by creating a request as a {semantic description}, like those being used and developed in \acp{SOA}}
\acrodef{picUserReq}{A population is then instantiated in the user's habitat in response to the user's request, seeded from the agents available at their habitat.}
\acrodef{statesCap}{possible evolutionary path through the state-space $I$}
\acrodef{capStates3}{the {selection pressure} of the evolutionary process}
\acrodef{capStates}{driving it towards the {maximal state} of the {maximum macro-state} $M_{max}$}
\acrodef{graphCap}{in the {maximum macro-state} $M_{max}$ only after generation 178 and always after generation 482.}
\acrodef{aScap}{With the mutation rate under or equal to 60\%, the evolving agent population showed no instability}
\acrodef{digEco}{with the agents, the populations, the agent migration for \acl{DEC}, and the environmental selection pressures provided by the user base, then the union of the habitats creates the Digital Ecosystem}
\acrodef{archComTop}{many strongly connected clusters (communities), called {sub-networks} (quasi-complete graphs), with a few connections between these clusters (communities). Graphs with this topology have a very high clustering coefficient and small characteristic path lengths}
\acrodef{similarCap}{requests are evaluated on separate {islands} (populations), and so adaptation is accelerated by the sharing of solutions between evolving populations (islands), because they are working to solve similar requests (problems).}
\acrodef{picUser}{will formulate queries to the Digital Ecosystem by creating a request as a {semantic description}, like those being used and developed in \acp{SOA}}
\acrodef{picUserReq}{A population is then instantiated in the user's habitat in response to the user's request, seeded from the agents available at their habitat.}
\acrodef{statesCap}{possible evolutionary path through the state-space $I$}
\acrodef{capStates3}{the {selection pressure} of the evolutionary process}
\acrodef{capStates}{driving it towards the {maximal state} of the {maximum macro-state} $M_{max}$}
\acrodef{graphCap}{in the {maximum macro-state} $M_{max}$ only after generation 178 and always after generation 482.}
\acrodef{aScap}{With the mutation rate under or equal to 60\%, the evolving agent population showed no instability}
\acrodef{digEco}{with the agents, the populations, the agent migration for \acl{DEC}, and the environmental selection pressures provided by the user base, then the union of the habitats creates the Digital Ecosystem}
\acrodef{archComTop}{many strongly connected clusters (communities), called {sub-networks} (quasi-complete graphs), with a few connections between these clusters (communities). Graphs with this topology have a very high clustering coefficient and small characteristic path lengths}
\acrodef{similarCap}{requests are evaluated on separate {islands} (populations), and so adaptation is accelerated by the sharing of solutions between evolving populations (islands), because they are working to solve similar requests (problems).}
\acrodef{picUser}{will formulate queries to the Digital Ecosystem by creating a request as a {semantic description}, like those being used and developed in \acp{SOA}}
\acrodef{picUserReq}{A population is then instantiated in the user's habitat in response to the user's request, seeded from the agents available at their habitat.}
\acrodef{statesCap}{possible evolutionary path through the state-space $I$}
\acrodef{capStates3}{the {selection pressure} of the evolutionary process}
\acrodef{capStates}{driving it towards the {maximal state} of the {maximum macro-state} $M_{max}$}
\acrodef{graphCap}{in the {maximum macro-state} $M_{max}$ only after generation 178 and always after generation 482.}
\acrodef{aScap}{With the mutation rate under or equal to 60\%, the evolving agent population showed no instability}
\acrodef{digEco}{with the agents, the populations, the agent migration for \acl{DEC}, and the environmental selection pressures provided by the user base, then the union of the habitats creates the Digital Ecosystem}
\acrodef{archComTop}{many strongly connected clusters (communities), called {sub-networks} (quasi-complete graphs), with a few connections between these clusters (communities). Graphs with this topology have a very high clustering coefficient and small characteristic path lengths}
\acrodef{similarCap}{requests are evaluated on separate {islands} (populations), and so adaptation is accelerated by the sharing of solutions between evolving populations (islands), because they are working to solve similar requests (problems).}
\acrodef{picUser}{will formulate queries to the Digital Ecosystem by creating a request as a {semantic description}, like those being used and developed in \acp{SOA}}
\acrodef{picUserReq}{A population is then instantiated in the user's habitat in response to the user's request, seeded from the agents available at their habitat.}
\acrodef{statesCap}{possible evolutionary path through the state-space $I$}
\acrodef{capStates3}{the {selection pressure} of the evolutionary process}
\acrodef{capStates}{driving it towards the {maximal state} of the {maximum macro-state} $M_{max}$}
\acrodef{graphCap}{in the {maximum macro-state} $M_{max}$ only after generation 178 and always after generation 482.}
\acrodef{aScap}{With the mutation rate under or equal to 60\%, the evolving agent population showed no instability}
\acrodef{digEco}{with the agents, the populations, the agent migration for \acl{DEC}, and the environmental selection pressures provided by the user base, then the union of the habitats creates the Digital Ecosystem}
\acrodef{archComTop}{many strongly connected clusters (communities), called {sub-networks} (quasi-complete graphs), with a few connections between these clusters (communities). Graphs with this topology have a very high clustering coefficient and small characteristic path lengths}
\acrodef{similarCap}{requests are evaluated on separate {islands} (populations), and so adaptation is accelerated by the sharing of solutions between evolving populations (islands), because they are working to solve similar requests (problems).}
\acrodef{picUser}{will formulate queries to the Digital Ecosystem by creating a request as a {semantic description}, like those being used and developed in \acp{SOA}}
\acrodef{picUserReq}{A population is then instantiated in the user's habitat in response to the user's request, seeded from the agents available at their habitat.}
\acrodef{statesCap}{possible evolutionary path through the state-space $I$}
\acrodef{capStates3}{the {selection pressure} of the evolutionary process}
\acrodef{capStates}{driving it towards the {maximal state} of the {maximum macro-state} $M_{max}$}
\acrodef{graphCap}{in the {maximum macro-state} $M_{max}$ only after generation 178 and always after generation 482.}
\acrodef{aScap}{With the mutation rate under or equal to 60\%, the evolving agent population showed no instability}
\acrodef{digEco}{with the agents, the populations, the agent migration for \acl{DEC}, and the environmental selection pressures provided by the user base, then the union of the habitats creates the Digital Ecosystem}
\acrodef{archComTop}{many strongly connected clusters (communities), called {sub-networks} (quasi-complete graphs), with a few connections between these clusters (communities). Graphs with this topology have a very high clustering coefficient and small characteristic path lengths}
\acrodef{similarCap}{requests are evaluated on separate {islands} (populations), and so adaptation is accelerated by the sharing of solutions between evolving populations (islands), because they are working to solve similar requests (problems).}
\acrodef{picUser}{will formulate queries to the Digital Ecosystem by creating a request as a {semantic description}, like those being used and developed in \acp{SOA}}
\acrodef{picUserReq}{A population is then instantiated in the user's habitat in response to the user's request, seeded from the agents available at their habitat.}
\acrodef{statesCap}{possible evolutionary path through the state-space $I$}
\acrodef{capStates3}{the {selection pressure} of the evolutionary process}
\acrodef{capStates}{driving it towards the {maximal state} of the {maximum macro-state} $M_{max}$}
\acrodef{graphCap}{in the {maximum macro-state} $M_{max}$ only after generation 178 and always after generation 482.}
\acrodef{aScap}{With the mutation rate under or equal to 60\%, the evolving agent population showed no instability}
\acrodef{digEco}{with the agents, the populations, the agent migration for \acl{DEC}, and the environmental selection pressures provided by the user base, then the union of the habitats creates the Digital Ecosystem}
\acrodef{archComTop}{many strongly connected clusters (communities), called {sub-networks} (quasi-complete graphs), with a few connections between these clusters (communities). Graphs with this topology have a very high clustering coefficient and small characteristic path lengths}
\acrodef{similarCap}{requests are evaluated on separate {islands} (populations), and so adaptation is accelerated by the sharing of solutions between evolving populations (islands), because they are working to solve similar requests (problems).}
\acrodef{picUser}{will formulate queries to the Digital Ecosystem by creating a request as a {semantic description}, like those being used and developed in \acp{SOA}}
\acrodef{picUserReq}{A population is then instantiated in the user's habitat in response to the user's request, seeded from the agents available at their habitat.}
\acrodef{statesCap}{possible evolutionary path through the state-space $I$}
\acrodef{capStates3}{the {selection pressure} of the evolutionary process}
\acrodef{capStates}{driving it towards the {maximal state} of the {maximum macro-state} $M_{max}$}
\acrodef{graphCap}{in the {maximum macro-state} $M_{max}$ only after generation 178 and always after generation 482.}
\acrodef{aScap}{With the mutation rate under or equal to 60\%, the evolving agent population showed no instability}
\acrodef{digEco}{with the agents, the populations, the agent migration for \acl{DEC}, and the environmental selection pressures provided by the user base, then the union of the habitats creates the Digital Ecosystem}
\acrodef{archComTop}{many strongly connected clusters (communities), called {sub-networks} (quasi-complete graphs), with a few connections between these clusters (communities). Graphs with this topology have a very high clustering coefficient and small characteristic path lengths}
\acrodef{similarCap}{requests are evaluated on separate {islands} (populations), and so adaptation is accelerated by the sharing of solutions between evolving populations (islands), because they are working to solve similar requests (problems).}
\acrodef{picUser}{will formulate queries to the Digital Ecosystem by creating a request as a {semantic description}, like those being used and developed in \acp{SOA}}
\acrodef{picUserReq}{A population is then instantiated in the user's habitat in response to the user's request, seeded from the agents available at their habitat.}
\acrodef{statesCap}{possible evolutionary path through the state-space $I$}
\acrodef{capStates3}{the {selection pressure} of the evolutionary process}
\acrodef{capStates}{driving it towards the {maximal state} of the {maximum macro-state} $M_{max}$}
\acrodef{graphCap}{in the {maximum macro-state} $M_{max}$ only after generation 178 and always after generation 482.}
\acrodef{aScap}{With the mutation rate under or equal to 60\%, the evolving agent population showed no instability}
\acrodef{digEco}{with the agents, the populations, the agent migration for \acl{DEC}, and the environmental selection pressures provided by the user base, then the union of the habitats creates the Digital Ecosystem}
\acrodef{archComTop}{many strongly connected clusters (communities), called {sub-networks} (quasi-complete graphs), with a few connections between these clusters (communities). Graphs with this topology have a very high clustering coefficient and small characteristic path lengths}
\acrodef{similarCap}{requests are evaluated on separate {islands} (populations), and so adaptation is accelerated by the sharing of solutions between evolving populations (islands), because they are working to solve similar requests (problems).}
\acrodef{picUser}{will formulate queries to the Digital Ecosystem by creating a request as a {semantic description}, like those being used and developed in \acp{SOA}}
\acrodef{picUserReq}{A population is then instantiated in the user's habitat in response to the user's request, seeded from the agents available at their habitat.}
\acrodef{statesCap}{possible evolutionary path through the state-space $I$}
\acrodef{capStates3}{the {selection pressure} of the evolutionary process}
\acrodef{capStates}{driving it towards the {maximal state} of the {maximum macro-state} $M_{max}$}
\acrodef{graphCap}{in the {maximum macro-state} $M_{max}$ only after generation 178 and always after generation 482.}
\acrodef{aScap}{With the mutation rate under or equal to 60\%, the evolving agent population showed no instability}
\acrodef{digEco}{with the agents, the populations, the agent migration for \acl{DEC}, and the environmental selection pressures provided by the user base, then the union of the habitats creates the Digital Ecosystem}
\acrodef{archComTop}{many strongly connected clusters (communities), called {sub-networks} (quasi-complete graphs), with a few connections between these clusters (communities). Graphs with this topology have a very high clustering coefficient and small characteristic path lengths}
\acrodef{similarCap}{requests are evaluated on separate {islands} (populations), and so adaptation is accelerated by the sharing of solutions between evolving populations (islands), because they are working to solve similar requests (problems).}
\acrodef{picUser}{will formulate queries to the Digital Ecosystem by creating a request as a {semantic description}, like those being used and developed in \acp{SOA}}
\acrodef{picUserReq}{A population is then instantiated in the user's habitat in response to the user's request, seeded from the agents available at their habitat.}
\acrodef{statesCap}{possible evolutionary path through the state-space $I$}
\acrodef{capStates3}{the {selection pressure} of the evolutionary process}
\acrodef{capStates}{driving it towards the {maximal state} of the {maximum macro-state} $M_{max}$}
\acrodef{graphCap}{in the {maximum macro-state} $M_{max}$ only after generation 178 and always after generation 482.}
\acrodef{aScap}{With the mutation rate under or equal to 60\%, the evolving agent population showed no instability}
\acrodef{digEco}{with the agents, the populations, the agent migration for \acl{DEC}, and the environmental selection pressures provided by the user base, then the union of the habitats creates the Digital Ecosystem}
\acrodef{archComTop}{many strongly connected clusters (communities), called {sub-networks} (quasi-complete graphs), with a few connections between these clusters (communities). Graphs with this topology have a very high clustering coefficient and small characteristic path lengths}
\acrodef{similarCap}{requests are evaluated on separate {islands} (populations), and so adaptation is accelerated by the sharing of solutions between evolving populations (islands), because they are working to solve similar requests (problems).}
\acrodef{picUser}{will formulate queries to the Digital Ecosystem by creating a request as a {semantic description}, like those being used and developed in \acp{SOA}}
\acrodef{picUserReq}{A population is then instantiated in the user's habitat in response to the user's request, seeded from the agents available at their habitat.}
\acrodef{statesCap}{possible evolutionary path through the state-space $I$}
\acrodef{capStates3}{the {selection pressure} of the evolutionary process}
\acrodef{capStates}{driving it towards the {maximal state} of the {maximum macro-state} $M_{max}$}
\acrodef{graphCap}{in the {maximum macro-state} $M_{max}$ only after generation 178 and always after generation 482.}
\acrodef{aScap}{With the mutation rate under or equal to 60\%, the evolving agent population showed no instability}
\acrodef{digEco}{with the agents, the populations, the agent migration for \acl{DEC}, and the environmental selection pressures provided by the user base, then the union of the habitats creates the Digital Ecosystem}
\acrodef{archComTop}{many strongly connected clusters (communities), called {sub-networks} (quasi-complete graphs), with a few connections between these clusters (communities). Graphs with this topology have a very high clustering coefficient and small characteristic path lengths}
\acrodef{similarCap}{requests are evaluated on separate {islands} (populations), and so adaptation is accelerated by the sharing of solutions between evolving populations (islands), because they are working to solve similar requests (problems).}
\acrodef{picUser}{will formulate queries to the Digital Ecosystem by creating a request as a {semantic description}, like those being used and developed in \acp{SOA}}
\acrodef{picUserReq}{A population is then instantiated in the user's habitat in response to the user's request, seeded from the agents available at their habitat.}
\acrodef{statesCap}{possible evolutionary path through the state-space $I$}
\acrodef{capStates3}{the {selection pressure} of the evolutionary process}
\acrodef{capStates}{driving it towards the {maximal state} of the {maximum macro-state} $M_{max}$}
\acrodef{graphCap}{in the {maximum macro-state} $M_{max}$ only after generation 178 and always after generation 482.}
\acrodef{aScap}{With the mutation rate under or equal to 60\%, the evolving agent population showed no instability}
\acrodef{digEco}{with the agents, the populations, the agent migration for \acl{DEC}, and the environmental selection pressures provided by the user base, then the union of the habitats creates the Digital Ecosystem}
\acrodef{archComTop}{many strongly connected clusters (communities), called {sub-networks} (quasi-complete graphs), with a few connections between these clusters (communities). Graphs with this topology have a very high clustering coefficient and small characteristic path lengths}
\acrodef{similarCap}{requests are evaluated on separate {islands} (populations), and so adaptation is accelerated by the sharing of solutions between evolving populations (islands), because they are working to solve similar requests (problems).}
\acrodef{picUser}{will formulate queries to the Digital Ecosystem by creating a request as a {semantic description}, like those being used and developed in \acp{SOA}}
\acrodef{picUserReq}{A population is then instantiated in the user's habitat in response to the user's request, seeded from the agents available at their habitat.}
\acrodef{statesCap}{possible evolutionary path through the state-space $I$}
\acrodef{capStates3}{the {selection pressure} of the evolutionary process}
\acrodef{capStates}{driving it towards the {maximal state} of the {maximum macro-state} $M_{max}$}
\acrodef{graphCap}{in the {maximum macro-state} $M_{max}$ only after generation 178 and always after generation 482.}
\acrodef{aScap}{With the mutation rate under or equal to 60\%, the evolving agent population showed no instability}
\acrodef{digEco}{with the agents, the populations, the agent migration for \acl{DEC}, and the environmental selection pressures provided by the user base, then the union of the habitats creates the Digital Ecosystem}
\acrodef{archComTop}{many strongly connected clusters (communities), called {sub-networks} (quasi-complete graphs), with a few connections between these clusters (communities). Graphs with this topology have a very high clustering coefficient and small characteristic path lengths}
\acrodef{similarCap}{requests are evaluated on separate {islands} (populations), and so adaptation is accelerated by the sharing of solutions between evolving populations (islands), because they are working to solve similar requests (problems).}
\acrodef{picUser}{will formulate queries to the Digital Ecosystem by creating a request as a {semantic description}, like those being used and developed in \acp{SOA}}
\acrodef{picUserReq}{A population is then instantiated in the user's habitat in response to the user's request, seeded from the agents available at their habitat.}
\acrodef{statesCap}{possible evolutionary path through the state-space $I$}
\acrodef{capStates3}{the {selection pressure} of the evolutionary process}
\acrodef{capStates}{driving it towards the {maximal state} of the {maximum macro-state} $M_{max}$}
\acrodef{graphCap}{in the {maximum macro-state} $M_{max}$ only after generation 178 and always after generation 482.}
\acrodef{aScap}{With the mutation rate under or equal to 60\%, the evolving agent population showed no instability}
\acrodef{digEco}{with the agents, the populations, the agent migration for \acl{DEC}, and the environmental selection pressures provided by the user base, then the union of the habitats creates the Digital Ecosystem}
\acrodef{archComTop}{many strongly connected clusters (communities), called {sub-networks} (quasi-complete graphs), with a few connections between these clusters (communities). Graphs with this topology have a very high clustering coefficient and small characteristic path lengths}
\acrodef{similarCap}{requests are evaluated on separate {islands} (populations), and so adaptation is accelerated by the sharing of solutions between evolving populations (islands), because they are working to solve similar requests (problems).}
\acrodef{picUser}{will formulate queries to the Digital Ecosystem by creating a request as a {semantic description}, like those being used and developed in \acp{SOA}}
\acrodef{picUserReq}{A population is then instantiated in the user's habitat in response to the user's request, seeded from the agents available at their habitat.}
\acrodef{statesCap}{possible evolutionary path through the state-space $I$}
\acrodef{capStates3}{the {selection pressure} of the evolutionary process}
\acrodef{capStates}{driving it towards the {maximal state} of the {maximum macro-state} $M_{max}$}
\acrodef{graphCap}{in the {maximum macro-state} $M_{max}$ only after generation 178 and always after generation 482.}
\acrodef{aScap}{With the mutation rate under or equal to 60\%, the evolving agent population showed no instability}
\acrodef{digEco}{with the agents, the populations, the agent migration for \acl{DEC}, and the environmental selection pressures provided by the user base, then the union of the habitats creates the Digital Ecosystem}
\acrodef{archComTop}{many strongly connected clusters (communities), called {sub-networks} (quasi-complete graphs), with a few connections between these clusters (communities). Graphs with this topology have a very high clustering coefficient and small characteristic path lengths}
\acrodef{similarCap}{requests are evaluated on separate {islands} (populations), and so adaptation is accelerated by the sharing of solutions between evolving populations (islands), because they are working to solve similar requests (problems).}
\acrodef{picUser}{will formulate queries to the Digital Ecosystem by creating a request as a {semantic description}, like those being used and developed in \acp{SOA}}
\acrodef{picUserReq}{A population is then instantiated in the user's habitat in response to the user's request, seeded from the agents available at their habitat.}
\acrodef{statesCap}{possible evolutionary path through the state-space $I$}
\acrodef{capStates3}{the {selection pressure} of the evolutionary process}
\acrodef{capStates}{driving it towards the {maximal state} of the {maximum macro-state} $M_{max}$}
\acrodef{graphCap}{in the {maximum macro-state} $M_{max}$ only after generation 178 and always after generation 482.}
\acrodef{aScap}{With the mutation rate under or equal to 60\%, the evolving agent population showed no instability}
\acrodef{digEco}{with the agents, the populations, the agent migration for \acl{DEC}, and the environmental selection pressures provided by the user base, then the union of the habitats creates the Digital Ecosystem}
\acrodef{archComTop}{many strongly connected clusters (communities), called {sub-networks} (quasi-complete graphs), with a few connections between these clusters (communities). Graphs with this topology have a very high clustering coefficient and small characteristic path lengths}
\acrodef{similarCap}{requests are evaluated on separate {islands} (populations), and so adaptation is accelerated by the sharing of solutions between evolving populations (islands), because they are working to solve similar requests (problems).}
\acrodef{picUser}{will formulate queries to the Digital Ecosystem by creating a request as a {semantic description}, like those being used and developed in \acp{SOA}}
\acrodef{picUserReq}{A population is then instantiated in the user's habitat in response to the user's request, seeded from the agents available at their habitat.}
\acrodef{statesCap}{possible evolutionary path through the state-space $I$}
\acrodef{capStates3}{the {selection pressure} of the evolutionary process}
\acrodef{capStates}{driving it towards the {maximal state} of the {maximum macro-state} $M_{max}$}
\acrodef{graphCap}{in the {maximum macro-state} $M_{max}$ only after generation 178 and always after generation 482.}
\acrodef{aScap}{With the mutation rate under or equal to 60\%, the evolving agent population showed no instability}
\acrodef{digEco}{with the agents, the populations, the agent migration for \acl{DEC}, and the environmental selection pressures provided by the user base, then the union of the habitats creates the Digital Ecosystem}
\acrodef{archComTop}{many strongly connected clusters (communities), called {sub-networks} (quasi-complete graphs), with a few connections between these clusters (communities). Graphs with this topology have a very high clustering coefficient and small characteristic path lengths}
\acrodef{similarCap}{requests are evaluated on separate {islands} (populations), and so adaptation is accelerated by the sharing of solutions between evolving populations (islands), because they are working to solve similar requests (problems).}
\acrodef{picUser}{will formulate queries to the Digital Ecosystem by creating a request as a {semantic description}, like those being used and developed in \acp{SOA}}
\acrodef{picUserReq}{A population is then instantiated in the user's habitat in response to the user's request, seeded from the agents available at their habitat.}
\acrodef{statesCap}{possible evolutionary path through the state-space $I$}
\acrodef{capStates3}{the {selection pressure} of the evolutionary process}
\acrodef{capStates}{driving it towards the {maximal state} of the {maximum macro-state} $M_{max}$}
\acrodef{graphCap}{in the {maximum macro-state} $M_{max}$ only after generation 178 and always after generation 482.}
\acrodef{aScap}{With the mutation rate under or equal to 60\%, the evolving agent population showed no instability}
\acrodef{digEco}{with the agents, the populations, the agent migration for \acl{DEC}, and the environmental selection pressures provided by the user base, then the union of the habitats creates the Digital Ecosystem}
\acrodef{archComTop}{many strongly connected clusters (communities), called {sub-networks} (quasi-complete graphs), with a few connections between these clusters (communities). Graphs with this topology have a very high clustering coefficient and small characteristic path lengths}
\acrodef{similarCap}{requests are evaluated on separate {islands} (populations), and so adaptation is accelerated by the sharing of solutions between evolving populations (islands), because they are working to solve similar requests (problems).}
\acrodef{picUser}{will formulate queries to the Digital Ecosystem by creating a request as a {semantic description}, like those being used and developed in \acp{SOA}}
\acrodef{picUserReq}{A population is then instantiated in the user's habitat in response to the user's request, seeded from the agents available at their habitat.}
\acrodef{statesCap}{possible evolutionary path through the state-space $I$}
\acrodef{capStates3}{the {selection pressure} of the evolutionary process}
\acrodef{capStates}{driving it towards the {maximal state} of the {maximum macro-state} $M_{max}$}
\acrodef{graphCap}{in the {maximum macro-state} $M_{max}$ only after generation 178 and always after generation 482.}
\acrodef{aScap}{With the mutation rate under or equal to 60\%, the evolving agent population showed no instability}
\acrodef{digEco}{with the agents, the populations, the agent migration for \acl{DEC}, and the environmental selection pressures provided by the user base, then the union of the habitats creates the Digital Ecosystem}
\acrodef{archComTop}{many strongly connected clusters (communities), called {sub-networks} (quasi-complete graphs), with a few connections between these clusters (communities). Graphs with this topology have a very high clustering coefficient and small characteristic path lengths}
\acrodef{similarCap}{requests are evaluated on separate {islands} (populations), and so adaptation is accelerated by the sharing of solutions between evolving populations (islands), because they are working to solve similar requests (problems).}
\acrodef{picUser}{will formulate queries to the Digital Ecosystem by creating a request as a {semantic description}, like those being used and developed in \acp{SOA}}
\acrodef{picUserReq}{A population is then instantiated in the user's habitat in response to the user's request, seeded from the agents available at their habitat.}
\acrodef{statesCap}{possible evolutionary path through the state-space $I$}
\acrodef{capStates3}{the {selection pressure} of the evolutionary process}
\acrodef{capStates}{driving it towards the {maximal state} of the {maximum macro-state} $M_{max}$}
\acrodef{graphCap}{in the {maximum macro-state} $M_{max}$ only after generation 178 and always after generation 482.}
\acrodef{aScap}{With the mutation rate under or equal to 60\%, the evolving agent population showed no instability}
\acrodef{digEco}{with the agents, the populations, the agent migration for \acl{DEC}, and the environmental selection pressures provided by the user base, then the union of the habitats creates the Digital Ecosystem}
\acrodef{archComTop}{many strongly connected clusters (communities), called {sub-networks} (quasi-complete graphs), with a few connections between these clusters (communities). Graphs with this topology have a very high clustering coefficient and small characteristic path lengths}
\acrodef{similarCap}{requests are evaluated on separate {islands} (populations), and so adaptation is accelerated by the sharing of solutions between evolving populations (islands), because they are working to solve similar requests (problems).}
\acrodef{picUser}{will formulate queries to the Digital Ecosystem by creating a request as a {semantic description}, like those being used and developed in \acp{SOA}}
\acrodef{picUserReq}{A population is then instantiated in the user's habitat in response to the user's request, seeded from the agents available at their habitat.}
\acrodef{statesCap}{possible evolutionary path through the state-space $I$}
\acrodef{capStates3}{the {selection pressure} of the evolutionary process}
\acrodef{capStates}{driving it towards the {maximal state} of the {maximum macro-state} $M_{max}$}
\acrodef{graphCap}{in the {maximum macro-state} $M_{max}$ only after generation 178 and always after generation 482.}
\acrodef{aScap}{With the mutation rate under or equal to 60\%, the evolving agent population showed no instability}
\acrodef{digEco}{with the agents, the populations, the agent migration for \acl{DEC}, and the environmental selection pressures provided by the user base, then the union of the habitats creates the Digital Ecosystem}
\acrodef{archComTop}{many strongly connected clusters (communities), called {sub-networks} (quasi-complete graphs), with a few connections between these clusters (communities). Graphs with this topology have a very high clustering coefficient and small characteristic path lengths}
\acrodef{similarCap}{requests are evaluated on separate {islands} (populations), and so adaptation is accelerated by the sharing of solutions between evolving populations (islands), because they are working to solve similar requests (problems).}
\acrodef{picUser}{will formulate queries to the Digital Ecosystem by creating a request as a {semantic description}, like those being used and developed in \acp{SOA}}
\acrodef{picUserReq}{A population is then instantiated in the user's habitat in response to the user's request, seeded from the agents available at their habitat.}
\acrodef{statesCap}{possible evolutionary path through the state-space $I$}
\acrodef{capStates3}{the {selection pressure} of the evolutionary process}
\acrodef{capStates}{driving it towards the {maximal state} of the {maximum macro-state} $M_{max}$}
\acrodef{graphCap}{in the {maximum macro-state} $M_{max}$ only after generation 178 and always after generation 482.}
\acrodef{aScap}{With the mutation rate under or equal to 60\%, the evolving agent population showed no instability}

\newtheorem{definition}{Definition}
\newtheorem{theorem}{Theorem}

\newcommand{\be}{\begin{equation}}
\newcommand{\eeq}[1]{\label{#1}\end{equation}}
\newcommand{\bx}{{\bf X}}
\newcommand{\by}{{\bf Y}}
\newcommand{\bone}{{\bf M_{max}}}
\newcommand{\bfif}{{\bf M_{half}}}

\newcommand{\bxi}{\mbox{\boldmath $\xi$}}
\newcommand{\ra}{1, \ldots, n}

\newcommand{\normallinespacing}{\renewcommand{\baselinestretch}{1.5} \normalsize}

\newcommand{\narrowlinespacing}{\renewcommand{\baselinestretch}{1.0} \normalsize}

\begin{document}

\title{Digital Ecosystems:\\Stability of Evolving Agent Populations}

\numberofauthors{2} 
\author{
\alignauthor
Philippe De Wilde\\
	\affaddr{Intelligent Systems Lab}\\
	\affaddr{Department of Computer Science}\\
	\affaddr{Heriot Watt University}\\
	\affaddr{United Kingdom}\\
	\email{pdw@hw.ac.uk}
\alignauthor
Gerard Briscoe\\
	\affaddr{Digital Ecosystems Lab}\\
	\affaddr{Department of Media and Communications}\\
	\affaddr{London School of Economics}\\
	\affaddr{United Kingdom}\\
	\email{g.briscoe@lse.ac.uk}
}

\maketitle
\begin{abstract}
Stability is perhaps one of the most desirable features of any engineered system, given the importance of being able to predict its response to various environmental conditions prior to actual deployment. Engineered systems are becoming ever more complex, approaching the same levels of biological ecosystems, and so their stability becomes ever more important, but taking on more and more differential dynamics can make stability an ever more elusive property. The Chli-DeWilde definition of stability views a \acl{MAS} as a discrete time Markov chain with potentially unknown transition probabilities. With a \acl{MAS} being considered stable when its state, a stochastic process, has converged to an equilibrium distribution, because stability of a system can be understood intuitively as exhibiting bounded behaviour. We investigate an extension to include \acp{MAS} with \emph{evolutionary dynamics}, focusing on the evolving agent populations of our Digital Ecosystem. We then built upon this to construct an entropy-based definition for the \emph{degree of instability} (entropy of the limit probabilities), which was later used to perform a \emph{stability analysis}. The Digital Ecosystem is considered to investigate the stability of an evolving agent population through simulations, for which the results were consistent with the original Chli-DeWilde definition. 
\end{abstract}



\pagebreak\section{Introduction}

Digital Ecosystems are distributed adaptive open socio-technical systems, with properties of self-organisation, scalability and sustainability, inspired by natural ecosystems \cite{thesis}, and are emerging as a novel approach to the catalysis of sustainable regional development driven by \acp{SME}. Digital Ecosystems aim to help local economic actors become active players in globalisation, valorising their local culture and vocations, and enabling them to interact and create value networks at the global level \cite{dini2008bid}. With its technical component being the \emph{digital} counterpart of a biological ecosystem, providing for the evolution of software services (agents) in a distributed network \cite{de07oz, dbebkpub}.

\emph{\acfp{MAS}} is a growing field, primarily because of recent developments of the Internet as a means of circulating information, goods and services; and many researchers have contributed valuable work in the area in recent years \cite{moaspaper}. However, despite both \emph{evolutionary computing} and \acp{MAS} being mature research areas \cite{masOverviewPaper, ecpaper} their integration is a recent development \cite{smith1998fec} and non-trivial, because agents can be modelled as state machines and \emph{evolutionary computing} algorithms have been developed to work on numerical data and strings without memory effects. Their integration also allows for the creation of the \emph{digital} counterpart of a biological ecosystem, what we call a Digital Ecosystem \cite{bionetics}, which provides a conceptual architecture for the evolution of software agents (services) in a distributed network \cite{javaOne}. Our aim here is to determine, for our Digital Ecosystem and other \acp{MAS} which make use of evolutionary computing \cite{mabu2007gbe, smith2000eec}, macroscopic variables that characterise the stability of their evolving agent populations and therefore the system as a whole.

\section{The Digital Ecosystem}

Our Digital Ecosystem \cite{bionetics} provides a two-level optimisation scheme inspired by natural ecosystems, in which a decentralised peer-to-peer network forms an underlying tier of distributed agents. These agents then feed a second optimisation level based on an evolutionary algorithm that operates locally on single habitats (peers), aiming to find solutions that satisfy locally relevant constraints. The local search is sped up through this twofold process, providing better local optima as the distributed optimisation provides prior sampling of the search space by making use of computations already performed in other peers with similar constraints \cite{javaOne}. So, the Digital Ecosystem supports the automatic combining of numerous agents (which represent services), by their interaction in evolving populations to meet user requests for applications, in a scalable architecture of distributed interconnected habitats. The sharing of agents between habitats ensures the system is scalable, while maintaining a high evolutionary specialisation for each user. The network of interconnected habitats is equivalent to the \emph{abiotic} environment of biological ecosystems; combined \setCap{with the agents, the populations, the agent migration for \acl{DEC}, and the environmental selection pressures provided by the user base, then the union of the habitats creates the Digital Ecosystem}{digEco}, which is summarised in Figure \ref{architecture2}. The continuous and varying user requests for applications provide a dynamic evolutionary pressure on the applications (agent aggregations), which have to evolve to better fulfil those user requests, and without which there would be no driving force to the evolutionary self-organisation of the Digital Ecosystem.

\tfigure{width=3.33in}{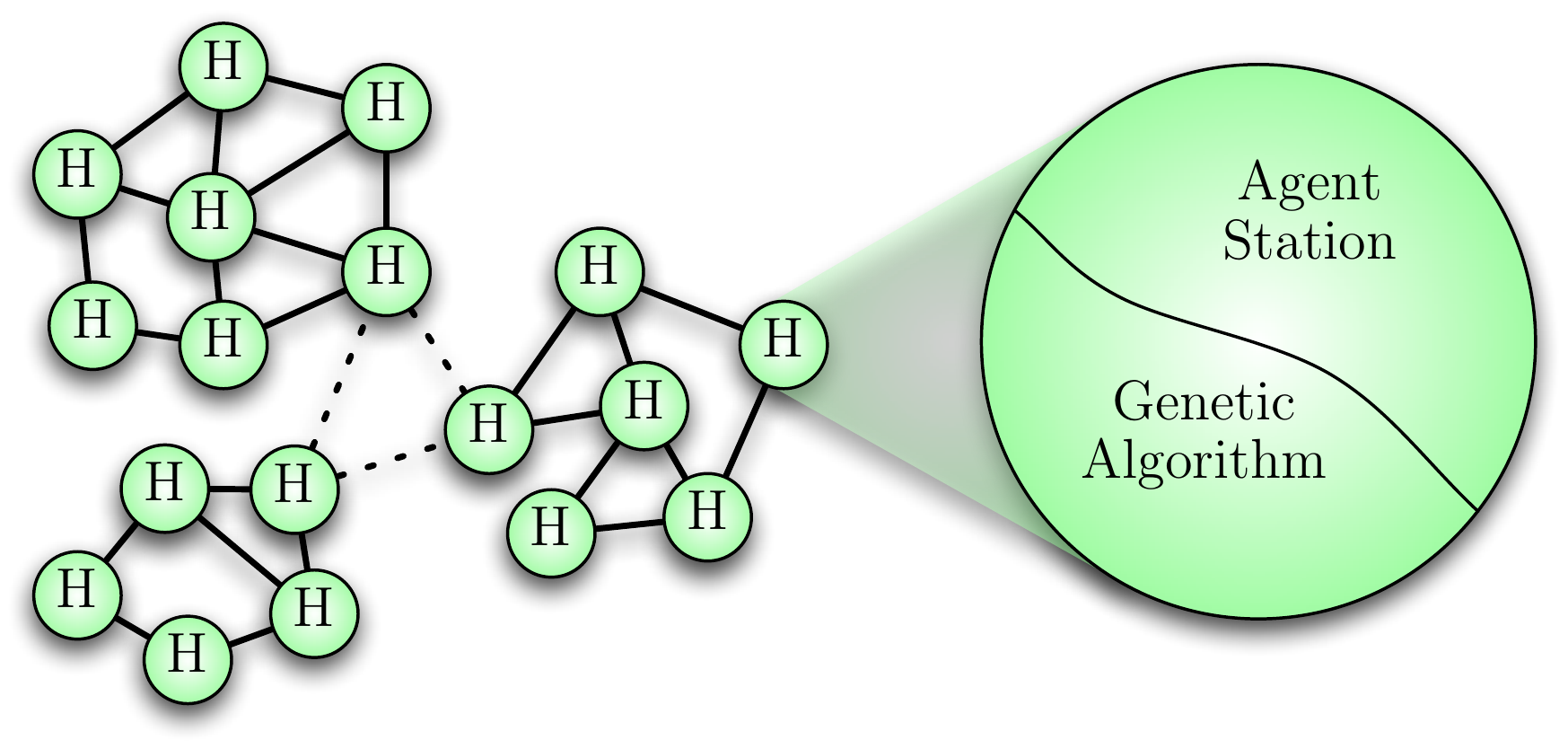}{graffle}{Digital Ecosystem}{Optimisation architecture in which agents (representing services) travel along the P2P connections; in every node (habitat) local optimisation is performed through an evolutionary algorithm, where the search space is determined by the agents present at the node.}{-9mm}{}{}{}

If we consider an example user base for the Digital Ecosystem, the use of \acp{SOA} in its definition means that \acf{B2B} interaction scenarios lend themselves to being a potential user base for Digital Ecosystems. So, we can consider a \emph{business ecosystem} of \acf{SME} networks \cite{moore1996}, as a specific class of examples for \ac{B2B} interaction scenarios; and in which the \ac{SME} users are requesting and providing software services, represented as agents in the Digital Ecosystem, to fulfil the needs of their business processes, creating a Digital Business Ecosystem as shown in Figure \ref{DBE}. \acp{SOA} promise to provide potentially huge numbers of services that programmers can combine, via the standardised interfaces, to create increasingly more sophisticated and distributed applications. The Digital Ecosystem extends this concept with the automatic combining of available and applicable services, represented by agents, in a scalable architecture, to meet user requests for applications. These agents will recombine and evolve over time, constantly seeking to improve their effectiveness for the user base. From the SME users' point of view the Digital Ecosystem provides a network infrastructure where connected enterprises can advertise and search for services (real-world or software only), putting a particular emphasis on the composability of loosely coupled services and their optimisation to local and regional, needs and conditions. To support these SME users the Digital Ecosystem is satisfying the companies' business requirements by finding the most suitable services or combination of services (applications) available in the network. An application (composition of services) is defined be an agent aggregation (collection) in the habitat network that can move from one peer (company) to another, being hosted only in those where it is most useful in satisfying the \ac{SME} users' business needs.

\tfigure{width=3.33in}{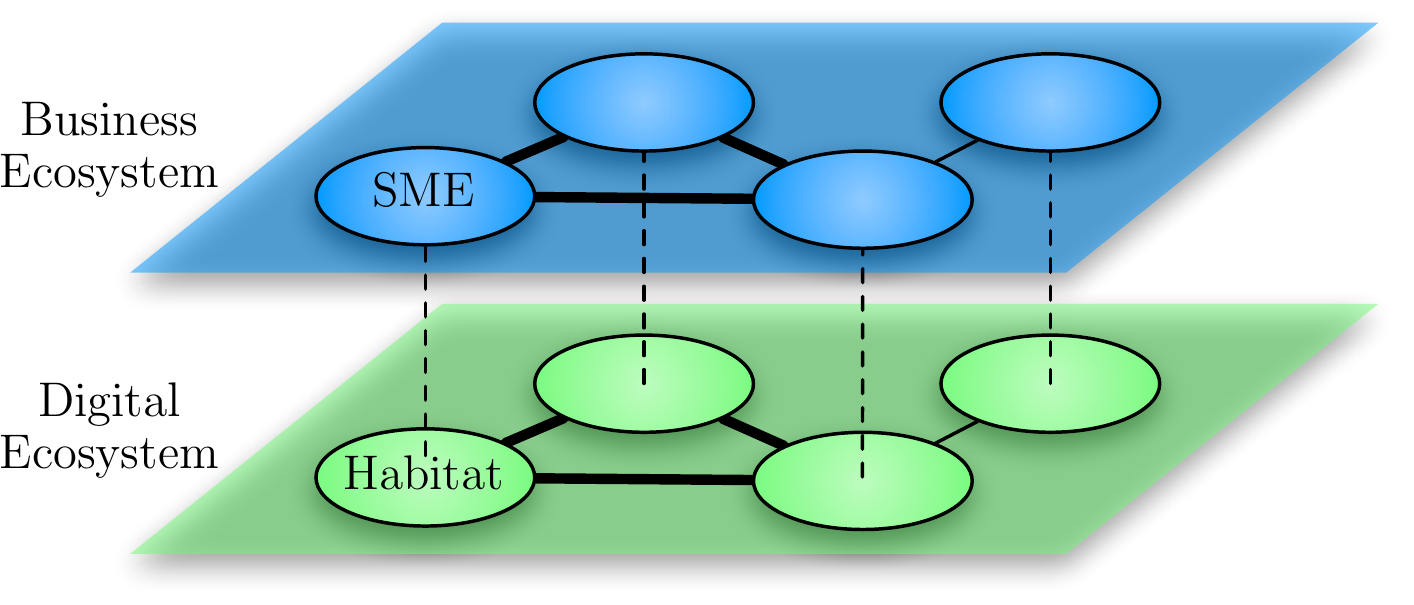}{graffle}{Digital Business Ecosystem}{Business ecosystem, network of \acp{SME} \cite{moore1996}, using the Digital Ecosystem. The habitat clustering will therefore be parallel to the business sector communities.}{-9mm}{}{}{}

The agents consist of an \emph{executable component} and an \emph{ontological description}. So, the Digital Ecosystem can be considered a \ac{MAS} which uses \emph{distributed evolutionary computing} to combine suitable agents in order to meet user requests for applications. 

The landscape, in energy-centric biological ecosystems, defines the connectivity between habitats. Connectivity of nodes in the digital world is generally not defined by geography or spatial proximity, but by information or semantic proximity. For example, connectivity in a peer-to-peer network is based primarily on bandwidth and information content, and not geography. The island-models of \acl{DEC} use an information-centric model for the connectivity of nodes (\emph{islands}) \cite{lin1994cgp}. However, because it is generally defined for one-time use (to evolve a solution to one problem and then stop) it usually has a fixed connectivity between the nodes, and therefore a fixed topology. So, supporting evolution in the Digital Ecosystem, with a multi-objective \emph{selection pressure} (fitness landscape with many peaks), requires a re-configurable network topology, such that habitat connectivity can be dynamically adapted based on the observed migration paths of the agents between the users within the habitat network. Based on the island-models of \acl{DEC} \cite{lin1994cgp}, each connection between the habitats is bi-directional and there is a probability associated with moving in either direction across the connection, with the connection probabilities affecting the rate of migration of the agents. However, additionally, the connection probabilities will be updated by the success or failure of agent migration using the concept of Hebbian learning: the habitats which do not successfully exchange agents will become less strongly connected, and the habitats which do successfully exchange agents will achieve stronger connections. This leads to a topology that adapts over time, resulting in a network that supports and resembles the connectivity of the user base. If we consider a \emph{business ecosystem}, network of \acp{SME}, as an example user base; such business networks are typically small-world networks. They have \setCap{many strongly connected clusters (communities), called \emph{sub-networks} (quasi-complete graphs), with a few connections between these clusters (communities). Graphs with this topology have a very high clustering coefficient and small characteristic path lengths}{archComTop} \cite{swn1}. So, the Digital Ecosystem will take on a topology similar to that of the user base, as shown in Figure \ref{DBE}.

The novelty of our approach comes from the evolving populations being created in response to \emph{similar} requests. So whereas in the island-models of \acl{DEC} there are multiple evolving populations in response to one request \cite{lin1994cgp}, here there are multiple evolving populations in response to \emph{similar} requests. In our Digital Ecosystems different \setCap{requests are evaluated on separate \emph{islands} (populations), and so adaptation is accelerated by the sharing of solutions between evolving populations (islands), because they are working to solve similar requests (problems).}{similarCap}

The users \setCap{will formulate queries to the Digital Ecosystem by creating a request as a \emph{semantic description}, like those being used and developed in \acp{SOA}}{picUser}, specifying an application they desire and submitting it to their local peer (habitat). This description defines a metric for evaluating the \emph{fitness} of a composition of agents, as a distance function between the \emph{semantic description} of the request and the agents' \emph{ontological descriptions}. \setCap{A population is then instantiated in the user's habitat in response to the user's request, seeded from the agents available at their habitat.}{picUserReq} This allows the evolutionary optimisation to be accelerated in the following three ways: first, the habitat network provides a subset of the agents available globally, which is localised to the specific user it represents; second, making use of applications (agent aggregations) previously evolved in response to the user's earlier requests; and third, taking advantage of relevant applications evolved elsewhere in response to similar requests by other users. The population then proceeds to evolve the optimal application (agent aggregation) that fulfils the user request, and as the agents are the base unit for evolution, it searches the available agent combination space. For an evolved agent aggregation (application) that is executed by the user, it then migrates to other peers (habitats) becoming hosted where it is useful, to combine with other agents in other populations to assist in responding to other user requests for applications.

\section{Agent Stability}

While there are several definitions of stability \cite{moreau2005sms, weiss1999msm, olfatisaber2007cac} defined for \acp{MAS}, they are not applicable primarily because of the evolutionary dynamics inherent in the context of evolving agent populations. Whereas Chli-DeWilde stability of \acp{MAS} \cite{chli2} may be suitable, because it models \acp{MAS} as Markov chains, which are an established modelling approach in evolutionary computing \cite{rudolph1998fmc}. A \ac{MAS} is viewed as a discrete time Markov chain with potentially unknown transition probabilities, in which the agents are modelled as Markov processes, and is considered to be \emph{stable} when its state, a stochastic process, has converged to an equilibrium distribution \cite{chli2}. Also, while there has been past work on modelling \emph{evolutionary computing} algorithms as Markov chains \cite{rudolph, nix, goldberg2, eibenAarts}, we have found none including \acp{MAS} despite both being mature research areas \cite{masOverviewPaper, ecpaper}, because their integration is a recent development \cite{smith1998fec}. We therefore decided to extend the existing Chli-DeWilde definition of agent stability to include the necessary \emph{evolutionary dynamics}.

Chil-DeWilde stability was created to provide a clear notion of stability in \acp{MAS} \cite{chli2}, because stability is perhaps one of the most desirable features of any engineered system, given the importance of being able to predict its response to various environmental conditions prior to actual deployment; and while computer scientists often talk about stable or unstable systems \cite{mspaper5ThomasSycara1998, mspaper9Balakrishnan1997}, they did so without having a concrete or uniform definition of stability. Also, other properties had been widely investigated, such as openness \cite{mspaperAbramov2001}, scalability \cite{mspaperMarwala2001} and adaptability \cite{mspaperSimoesMarques2003}, but stability had not. So, the Chli-DeWilde definition of stability for \acp{MAS} was created \cite{chli2}, based on the stationary distribution of a stochastic system, modelling the agents as Markov processes, and therefore viewing a \ac{MAS} as a discrete time Markov chain with a potentially unknown transition probability distribution. The \ac{MAS} is considered to be stable once its state, a stochastic process, has converged to an equilibrium distribution \cite{chli2}, because stability of a system can be understood intuitively as exhibiting bounded behaviour.

Chli-DeWilde stability was derived \cite{chlithesis} from the notion of stability defined by De Wilde \cite{mspaperDeWilde1999a, mspaperLee1998}, based on the stationary distribution of a stochastic system, making use of discrete-time Markov chains, which we will now introduce\arxivfootnote{A more comprehensive introduction to Markov chain theory and stochastic processes is available in \cite{msthesisNorris1997} and \cite{msthesisCoxMiller1972}.}. If we let $I$ be a \emph{countable set}, in which each $i \in I$ is called a \emph{state} and $I$ is called the \emph{state-space}. We can then say that $\lambda = (\lambda_i : i \in I)$ is a \emph{measure on} $I$ if $0 \le \lambda_i < \infty$ for all $i \in I$, and additionally a \emph{distribution} if $\sum_{i \in I}{\lambda_i=1}$ \cite{chlithesis}. So, if $X$ is a \emph{random variable} taking values in $I$ and we have $\lambda_i = \Pr(X = i)$, then $\lambda$ is \emph{the distribution of $X$}, and we can say that a matrix $P = (p_{ij} : i,j \in I)$ is \emph{stochastic} if every row $(p_{ij} : j \in I)$ is a \emph{distribution} \cite{chlithesis}. We can then extend familiar notions of matrix and vector multiplication to cover a general index set $I$ of potentially infinite size, by defining the multiplication of a matrix by a measure as $\lambda P$, which is given by
\begin{equation}
(\lambda P)_i = \sum\limits_{j \in I}{\lambda_{j}p_{ij}}.
\label{ms3dot1}
\end{equation}
We can now describe the rules for a Markov chain by a definition in terms of the corresponding matrix $P$ \cite{chlithesis}.\\

\begin{definition}
We say that $(X^t)_{t\ge0}$ is a Markov chain with initial distribution $\lambda = (\lambda_i : i \in I)$ and transition matrix $P = (p_{ij} : i,j \in I)$ if:
\narrowlinespacing
\begin{enumerate}
\item $\Pr(X^0 = i_0) = \lambda_{i_0}$ and
\item $\Pr(X^{t+1} = i_{t+1}\ |\ X^0 = i_0, \ldots, X^t = i_t) = p_{i_t i_{t+1}}$.
\end{enumerate}
\vspace{-3mm}
\normallinespacing
We abbreviate these two conditions by saying that $(X^t)_{t\ge0}$ is Markov$(\lambda, P)$.
\end{definition}

In this first definition the Markov process is \emph{memoryless}, resulting in only the current state of the system being required to describe its subsequent behaviour. We say that a Markov process $X^0, X^1, \ldots, X^t$ has a \emph{stationary distribution} if the probability distribution of $X^t$ becomes independent of the time $t$ \cite{chli2}. So, the following theorem is an \emph{easy consequence} of the second condition from the first definition.\\

\begin{theorem}
A discrete-time random process $(X^t)_{t\ge0}$ is Markov$(\lambda,P)$, if and only if for all $t$ and $i_0, \ldots, i_t$ we have
\narrowlinespacing
\begin{equation}
\Pr(X^0 = i_0, \ldots, X^t = i_t) = \lambda_{i_0}p_{i_0 i_1} \cdots p_{i_{t-1}i_t}.
\label{ms3dot2}
\end{equation}
\vspace{-6mm}
\normallinespacing
\end{theorem}
 
This first theorem depicts the structure of a Markov chain, illustrating the relation with the stochastic matrix $P$, and defining its time-invariance property \cite{chlithesis}.\\

\begin{theorem}
Let $(X^t)_{t\ge0}$ be $Markov(\lambda,P)$, then for all $t,s\ge0$:
\narrowlinespacing
\begin{enumerate}
\item $\Pr(X^t = j) = (\lambda P^t)_j$ and
\item $\Pr(X^t = j\ |\ X^0 = i) = \Pr(X^{t+s} = j\ |\ X^s = i) = (P^t)_{ij}$.
\end{enumerate}
\vspace{-3mm}
\normallinespacing
\label{ms3dot3dot2}
For convenience $(P^t)_{ij}$ can be more conveniently denoted as $p^{(t)}_{ij}$.
\end{theorem}

Given this second theorem we can define $p^{(t)}_{ij}$ as the t-step transition probability from the state $i$ to $j$ \cite{chlithesis}, and we can now introduce the concept of an \emph{invariant distribution} \cite{chlithesis}, in which we say that $\lambda$ is invariant if
\begin{equation}
\lambda P = \lambda .
\end{equation}
The next theorem will link the existence of an \emph{invariant distribution}, which is an algebraic property of the matrix $P$, with the probabilistic concept of an \emph{equilibrium distribution}. This only applies to a restricted class of Markov chains, namely those with \emph{irreducible} and \emph{aperiodic} stochastic matrices. However, there is a multitude of analogous results for other types of Markov chains to which we can refer \cite{msthesisNorris1997, msthesisCoxMiller1972}, and the following theorem is provided as an indication of the family of theorems that apply. An \emph{irreducible} matrix $P$ is one for which, for all $i,j \in I$ there are sufficiently large $t,p^{(t)}_{ij} > 0$, and is \emph{aperiodic} if for all states $i \in I$ we have $p^{(t)}_{ii} > 0$ for all sufficiently large $t$ \cite{chlithesis}.\\

\begin{theorem}
Let $P$ be irreducible, aperiodic and have an invariant distribution, $\lambda$ can be any distribution, and suppose that $(X^t)_{t\ge0}$ is Markov$(\lambda, P)$ \cite{chlithesis}, then
\narrowlinespacing
\begin{eqnarray}
& \Pr(X^t = j) \to p_{j}^\infty \ as\ t \to \infty\ \mbox{for all}\ j \in I & \\
& and & \nonumber \\
& p^{(t)}_{ij} \to p_{j}^\infty \ as\ t \to \infty\ \mbox{for all}\ i,j \in I. &
\end{eqnarray}
\vspace{-9mm}
\normallinespacing
\end{theorem} 
 
We can now view a system $S$ as a countable set of states $I$ with implicitly defined transitions $P$ between them, and at time $t$ the state of the system is the random variable $X^t$, with the key assumption that $(X^t)_{t,0}$ is Markov$(\lambda,P)$ \cite{chlithesis}.\\

\begin{definition}
The system $S$ is said to be stable when the distribution of the its states converge to an \emph{equilibrium distribution},
\narrowlinespacing
\begin{equation}
\Pr(X^t = j) \to p_{j}^\infty \ as\ t \to \infty\ for\ all j\ \in I.
\end{equation}
\vspace{-9mm}
\normallinespacing
\end{definition}

More intuitively, the system $S$, a stochastic process $X^0$,$X^1$,$X^2$,... is \emph{stable} if the probability distribution of $X^t$ becomes independent of the time index $t$ for large $t$ \cite{chli2}. Most Markov chains with a finite state-space and positive transition probabilities are examples of stable systems, because after an initialisation period they settle down on a stationary distribution \cite{chlithesis}.

A \ac{MAS} can be viewed as a system $S$, with the system state represented by a finite vector $\bx$, having dimensions large enough to manage the agents present in the system. The state vector will consist of one or more elements for each agent, and a number of elements to define general properties of the system state. We can then model an agent as being \emph{dead}, i.e. not being present in the system, by setting the vector elements for that agent to some predefined null value \cite{chlithesis}.

\subsection{Extensions for Evolving Populations}
\label{def}

Extending Chli-DeWilde stability to the \emph{class} of \acp{MAS} that make use of \emph{evolutionary computing} algorithms, including our evolving agent populations, requires consideration of the following issues: the inclusion of \emph{population dynamics}, and an understanding of population \emph{macro-states}.

\subsubsection{Population Dynamics}

First, the \ac{MAS} of an evolving agent population is composed of $n$ agent aggregations, with each agent aggregation $i$ in a state $\xi_i^t$ at time $t$, where $i=1, 2, \ldots, n$. The states of the agent aggregations are \emph{random variables}, and so the state vector for the \ac{MAS} is a vector of random variables $\bxi^t$, with the time being discrete, $t=0, 1, \ldots$ . The interactions among the agent aggregations are noisy, and are given by the probability distributions
\be
\Pr(X_i | \by) = \Pr(\xi_i^{t+1} = X_i | \bxi^t = \by) , \quad \ra,
\eeq{eq1}
where $X_i$ is a value for the state of agent aggregation $i$, and $\by$ is a value for the state vector of the \ac{MAS}. The probabilities implement a Markov process, with the noise caused by mutations. Furthermore, the agent aggregations are individually subjected to a \emph{selection pressure} from the environment of the system, which is applied equally to all the agent aggregations of the population. So, the probability distributions are statistically independent, and
\be
\Pr(\bx | \by) = \Pi_{i=1}^n \Pr(\xi_i^{t+1} = X_i | \bxi^t = \by).
\eeq{eq5}
If the occupation probability of state $\bx$ at time $t$ is denoted by $p_{\bx}^t$, then
\be
p_{\bx}^t = \sum_{\by} \Pr(\bx | \by) p_{\by}^{t-1}.
\eeq{eq5.1}
This is a discrete time equation used to calculate the evolution of the state occupation probabilities from $t=0$, while equation (\ref{eq5}) is the probability of moving from one state to another. The \ac{MAS} (evolving agent population) is self-stabilising if the limit distribution of the occupation probabilities exists and is non-uniform, i.e.
\be
p_\bx^\infty = lim_{t \rightarrow \infty} p_{\bx}^t
\eeq{eq2}
exists for all states $\bx$, and there exist states $\bx$ and $\by$ such that
\be
p_\bx^\infty \neq p_\by^\infty.
\eeq{eq3}
These equations define that some configurations of the system, after an extended time, will be more likely than others, because the likelihood of their occurrence no longer changes. Such a system is \emph{stable}, because the likelihood of states occurring no longer changes with time, and is the definition of stability developed in \cite{chli2}. While equation (\ref{eq2}) is the \emph{probabilistic equivalence} of an \emph{attractor}\footnote{An attractor is a set of states, invariant under the dynamics, towards which neighbouring states asymptotically approach during evolution.} in a system with deterministic interactions, which we had to extend to a stochastic process because mutation is inherent in evolutionary dynamics.

Although the number of agents in the Chli-DeWilde formalism can vary, we require it to vary according to the \emph{selection pressure} acting upon the evolving agent population. We must therefore formally define and extend the definition of \emph{dead} agents, by introducing a new state $d$ for each agent aggregation. If an agent aggregation is in this state, $\xi_i^t=d$, then it is \emph{dead} and does not affect the state of other agent aggregations in the population. If an agent aggregation $i$ has low fitness then that agent aggregation will likely die, because
\be
\Pr(d | \by) = \Pr(\xi_i^{t+1} = d | \bxi^t = \by)
\eeq{eq4}
will be high for all $\by$. Conversely, if an agent aggregation has high fitness, then it will likely replicate, assuming the state of a similarly successful agent aggregation (mutant), or crossover might occur changing the state of the successful agent aggregation and another agent aggregation.

\subsubsection{Population Macro-States}

As we defined earlier, the state of an evolving agent population is determined by the collection of agent aggregations of which it consists at a specific time $t$, and potentially changing state as the time $t$ increases. So, we can define a macro-state $M$ as a set of states with a common property, here possessing at least one copy of the \emph{current maximum fitness individual}. Therefore, by its definition, each macro-state $M$ must also have a \emph{maximal state} composed entirely of copies of the \emph{current maximum fitness individual}. There must also be a macro-state consisting of all the states that have at least one copy of the \emph{global maximum fitness individual}, which we will call the \emph{maximum macro-state} $M_{max}$.

\tfigure{width=3.33in}{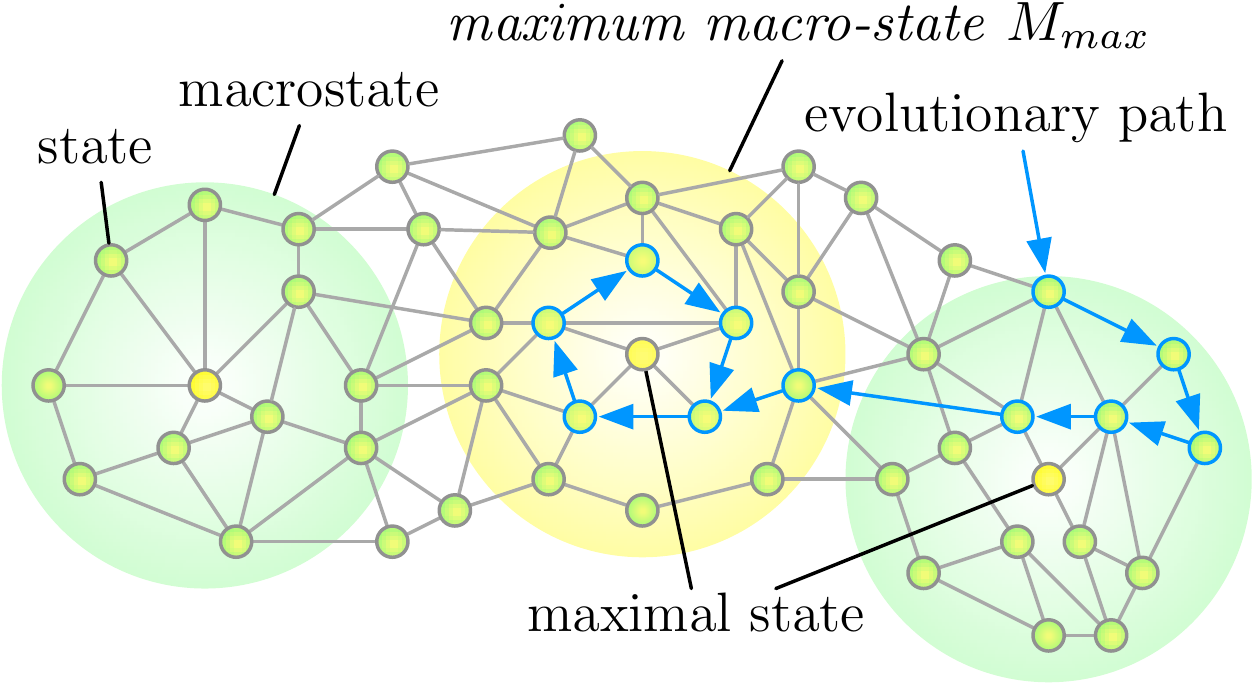}{graffle}{State-Space of an Evolving Agent Population}{A \getCap{statesCap} is shown, with \getCap{capStates3} \getCap{capStates}.}{-7mm}{!b}{}{}

We can consider the \emph{macro-states} of an evolving agent population visually through the representation of the state-space $I$ of the system $S$ shown in Figure \ref{states}, which includes a \setCap{possible evolutionary path through the state-space $I$}{statesCap}. Traversal through the state-space $I$ is directed by \setCap{the \emph{selection pressure} of the evolutionary process}{capStates3} acting upon the population $S$, \setCap{driving it towards the \emph{maximal state} of the \emph{maximum macro-state} $M_{max}$}{capStates}, which consists entirely of copies of the \emph{optimal solution}, and is the equilibrium state that the system $S$ is forever \emph{falling towards} without ever quite reaching, because of the noise (mutation) within the system. So, while this \emph{maximal state} will never be reached, the \emph{maximum macro-state} $M_{max}$ itself is certain to be reached, provided the system does not get trapped at local optima, i.e. the probability of being in the \emph{maximum macro-state} $M_{max}$ at infinite time is one, $p^{\infty}_{\bone}=1$, as defined from equation (\ref{eq5.1}).

Furthermore, we can define quantitatively the probability distribution of the macro-states that the system occupies at infinite time. For a stable system, as defined by equation (\ref{eq3}), the \emph{degree of instability}, $d_{ins}$, can be defined as the entropy of its probability distribution at infinite time,
\be
d_{ins} = H(p^\infty) = -\sum\limits_{\bx}p_{\bx}^{\infty}log_{N}(p_{\bx}^{\infty}),
\eeq{eq6}
where $N$ is the number of possible states, and taking $log$ to the base $N$ normalises the \emph{degree of instability}. The \emph{degree of instability} will range between zero (inclusive) and one (exclusive), because a maximum instability of one would only occur during the theoretical extreme scenario of a \emph{non-discriminating selection pressure}.

\section{Simulation and Results}

A simulated population of agent aggregations, $[A_1, A_1, A_2,$ $...]$, was evolved to solve user requests, seeded with agents from the \emph{agent-pool} of the habitats in which they were instantiated. A dynamic population size was used to ensure exploration of the available combinatorial search space, which increased with the average size of the population's agent aggregations. The optimal combination of agents (agent aggregation) was evolved to the user request $R$, by an artificial \emph{selection pressure} created by a \emph{fitness function} generated from the user request $R$. An individual (agent) of the population consisted of a set of attributes, ${a_1, a_2, ...}$, and a user request consisted of a set of required attributes, ${r_1, r_2, ...}$. So, the \emph{fitness function} for evaluating an individual agent aggregation $A$, relative to a user request $R$, was
\begin{equation}
fitness(A,R) = \frac{1}{1 + \sum_{r \in R}{|r-a|}},
\label{ff}
\end{equation}
where $a$ is the member of $A$ such that the difference to the required attribute $r$ was minimised. Equation \ref{ff} was used to assign \emph{fitness} values between 0.0 and 1.0 to each individual of the current generation of the population, directly affecting their ability to replicate into the next generation. The evolutionary computing process was encoded with a low mutation rate, a fixed selection pressure and a non-trapping fitness function (i.e. did not get trapped at local optima). The type of selection used \emph{fitness-proportional} and \emph{non-elitist}, \emph{fitness-proportional means that the \emph{fitter} the individual the higher its probability of} surviving to the next generation. \emph{Non-elitist} means that the best individual from one generation was not guaranteed to survive to the next generation; it had a high probability of surviving into the next generation, but it was not guaranteed as it might have been mutated. \emph{Crossover} (recombination) was then applied to a randomly chosen 10\% of the surviving population. \emph{Mutations} were then applied to a randomly chosen 10\% of the surviving population; \emph{point mutations} were randomly located, consisting of \emph{insertions} (an agent was inserted into an agent aggregation), \emph{replacements} (an agent was replaced in an agent aggregation), and \emph{deletions} (an agent was deleted from an agent aggregation). The issue of bloat was controlled by augmenting the \emph{fitness function} with a \emph{parsimony pressure} which biased the search to smaller agent aggregations, evaluating larger than average agent aggregations with a reduced \emph{fitness}, and thereby providing a dynamic control limit which adapted to the average size of the individuals of the ever-changing evolving agent populations.

Our evolving agent population (a \ac{MAS} with evolutionary dynamics) is stable if the distribution of the limit probabilities exists and is non-uniform, as defined by equations (\ref{eq2}) and (\ref{eq3}). The simplest case is a typical evolving agent population with one global optimal solution, which is stable if there are at least two macro-states with different limit occupation probabilities. We shall consider the \emph{maximum macro-state} $M_{max}$ and the \emph{sub-optimal macro-state} $M_{half}$. Where the states of the macro-state $M_{max}$ each possess at least one individual with global maximum fitness,
\begin{equation*}
p_{\bone}^\infty = lim_{t \rightarrow \infty} p_{\bone}^{(t)} = 1,
\end{equation*}
while the states of the macro-state $M_{half}$ each possess at least one individual with a fitness equal to \emph{half} of the global maximum fitness,
\begin{equation*}
p_{\bfif}^\infty = lim_{t \rightarrow \infty} p_{\bfif}^{(t)} = 0,
\end{equation*}
thereby fulfilling the requirements of equations (\ref{eq2}) and (\ref{eq3}). The \emph{sub-optimal macro-state} $M_{half}$, having a lower fitness, is predicted to be seen earlier in the evolutionary process before disappearing as higher fitness macro-states are reached. The system $S$ will take longer to reach the \emph{maximum macro-state} $M_{max}$, but once it does will likely remain, leaving only briefly depending on the strength of the mutation rate, as the \emph{selection pressure} is \emph{non-elitist}.

\tfigure{width=3.33in}{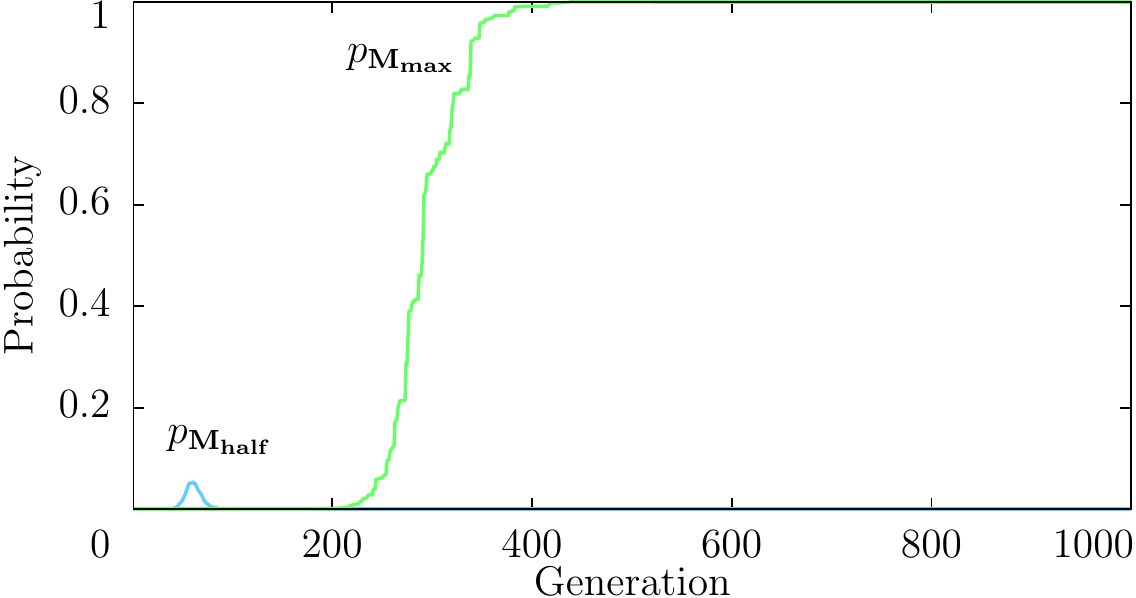}{graph}{Graph of the Probabilities of the Macro-States}{$M_{max}$ and $M_{half}$ at each Generation: The system $S$, a typical evolving agent population, was \getCap{graphCap}}{-7mm}{}{}{}

A value of $t=1000$ was chosen to represent $t=\infty$ experimentally, because the simulation has often been observed to reach the \emph{maximum macro-state} $M_{max}$ within 500 generations. Therefore, the probability of the system $S$ being in the \emph{maximum macro-state} $M_{max}$ at the thousandth generation is expected to be one, $p^{1000}_{\bone} = 1$. Furthermore, the probability of the system being in the \emph{sub-optimal macro-state} $M_{half}$ at the thousandth generation is expected to be zero, $p^{1000}_{\bfif} = 0$.

Figure \ref{macrostates} shows, for a typical evolving agent population, a graph of the probability as defined by equation (\ref{eq5.1}) of the \emph{maximum macro-state} $M_{max}$ and the \emph{sub-optimal macro-state} $M_{half}$ at each generation, averaged from ten thousand simulation runs for statistical significance. The behaviour of the simulated system $S$ was as expected, being \setCap{in the \emph{maximum macro-state} $M_{max}$ only after generation 178 and always after generation 482.}{graphCap} It was also observed being in the \emph{sub-optimal macro-state} $M_{half}$ only between generations 37 and 113, with a maximum probability of 0.053 (3 d.p.) at generation 61, and was such because the evolutionary path (state transitions) could avoid visiting the macro-state. As expected the probability of being in the \emph{maximum macro-state} $M_{max}$ at the thousandth generation was one, $p^{1000}_{\bone} = 1$, and so the probability of being in any other macro-state, including the \emph{sub-optimal macro-state} $M_{half}$, at the thousandth generation was zero, $p^{1000}_{\bfif} = 0$.

\tfigure{width=3.33in}{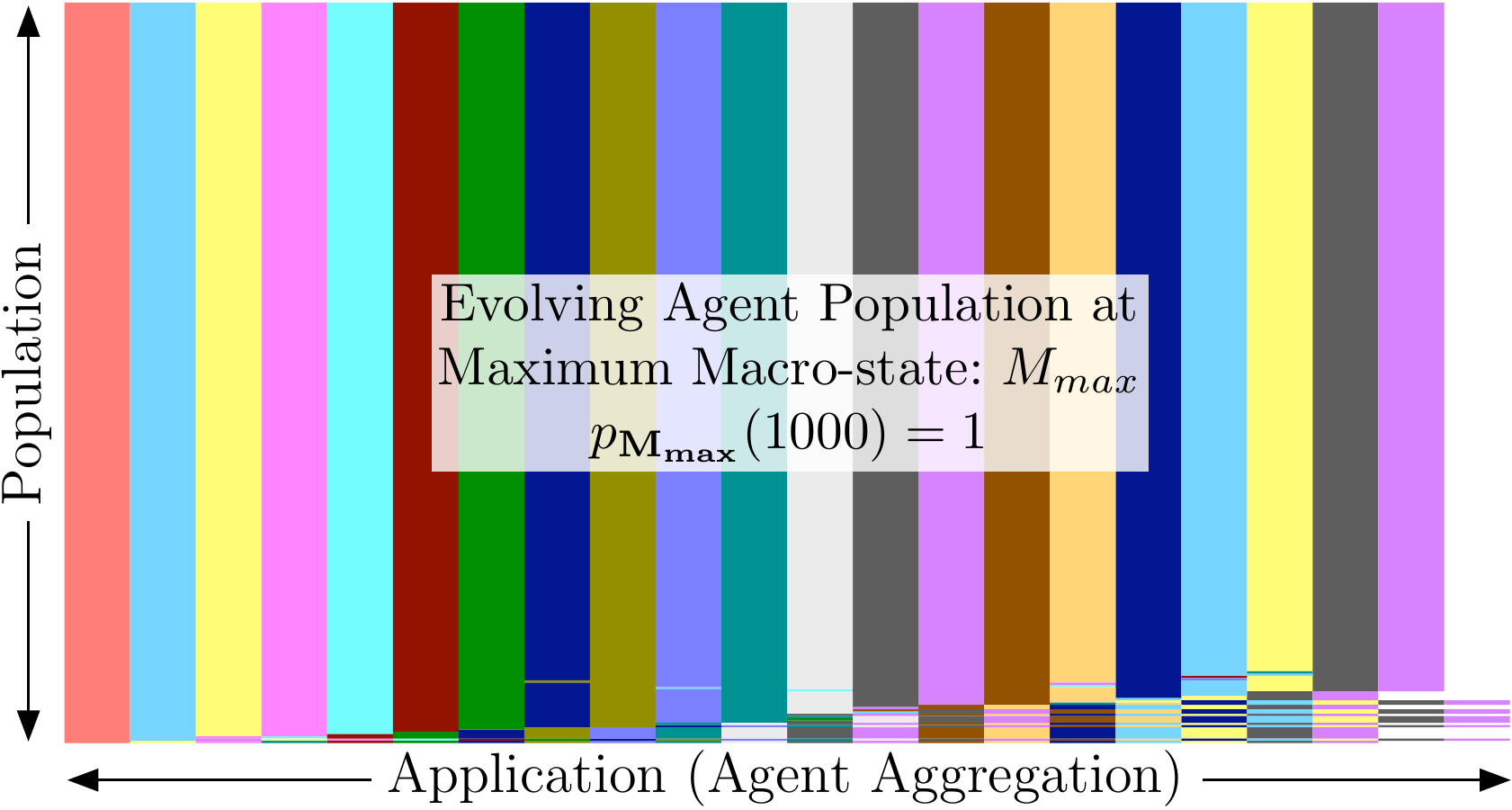}{graffle}{Visualisation of an Evolving Agent Population at the 1000th Generation}{The population consists of multiple agent aggregations, with each line representing an agent aggregation, and therefore each colour representing an agent.}{-8mm}{}{}{}

A visualisation for the state of a typical evolving agent population at the thousandth generation is shown in Figure \ref{vis62}, with each line representing an agent aggregation and each colour representing an agent, with the identical agent aggregations grouped for clarity. It shows that the evolving agent population reached the \emph{maximum macro-state} $M_{max}$ and remained there, but as expected never reached the \emph{maximal state} of the \emph{maximum macro-state}, where all the agent aggregations are identical and have maximum fitness, which is indicated by the lack of total uniformity in Figure \ref{vis62}. This was expected, because of the mutation (noise) within the evolutionary process, which is necessary to create the opportunity to find fitter (better) sequences and potentially avoid getting trapped at any local optima that may be present.

\subsection{Degree of Instability}

Given that our simulated evolving agent population is stable as defined by equations (\ref{eq2}) and (\ref{eq3}), we can determine the \emph{degree of instability} as defined by equation (\ref{eq6}). So, calculated from its limit probabilities, the \emph{degree of instability} was

\vspace{-5mm}

\begin{eqnarray}
d_{ins} = H(p^{1000}) &=& -\sum\limits_{\bx}p^{1000}_\bx log_{N}(p^{1000}_\bx) \nonumber \\ 
 &=& -1log_{N}(1) \nonumber \\
 &=& 0, \nonumber
\label{result2}
\end{eqnarray}

where $t=1000$ is an effective estimate for $t=\infty$, as explained earlier. The result was as expected because the \emph{maximum macro-state} $M_{max}$ at the thousandth generation was one, $p^{1000}_{\bone} = 1$, and so the probability of being in the other macro-states at the thousandth generation was zero. The system therefore shows no instability, as there is no entropy in the occupied macro-states at infinite time.

\subsection{Stability Analysis}

We then performed a \emph{stability analysis} (similar to a \emph{sensitivity analysis}) of a typical evolving agent population, varying key parameters within the simulation. We varied the mutation and crossover rates from 0\% to 100\% in 10\% increments, calculating the \emph{degree of instability}, $d_{ins}$ from (\ref{eq6}), at the thousandth generation. These \emph{degree of instability} values were averaged over ten thousand simulation runs, and graphed against the mutation and crossover rates in Figure \ref{stabilityAnalysis}. It shows that the crossover rate had little effect on the stability of our simulated evolving agent population, whereas the mutation rate did significantly affect the stability. \setCap{With the mutation rate under or equal to 60\%, the evolving agent population showed no instability}{aScap}, with $d_{ins}$ values equal to zero as the system $S$ was always in the same macro-state $M$ at infinite time, independent of the crossover rate. With the mutation rate above 60\% the instability increased significantly, with the system being in one of several different macro-states at infinite time; with a mutation rate of 70\% the system was still very stable, having low $d_{ins}$ values ranging between 0.08 and 0.16, but once the mutation rate was 80\% or greater the system became quite unstable, shown by high $d_{ins}$ values nearing 0.5. 

\tfigure{width=3.33in}{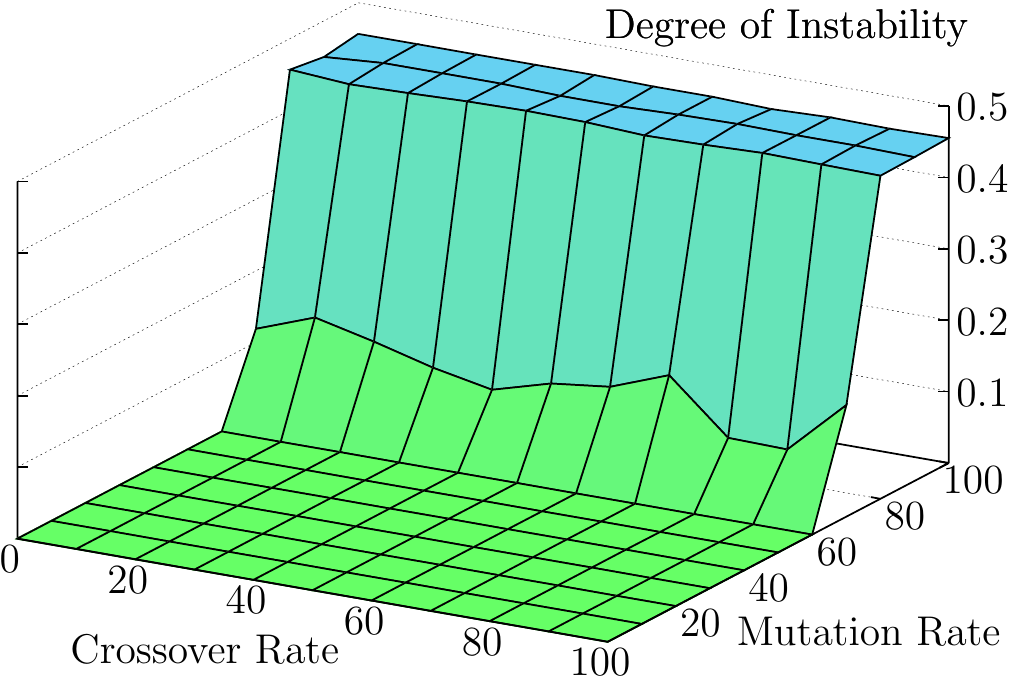}{graph}{Graph of Stability with Different Mutation and Crossover Rates}{\getCap{aScap}.}{-7mm}{!b}{}{}

As one would have expected, an \emph{extremely} high mutation rate has a destabilising effect on the \emph{stability} of an evolving agent population. The crossover rate had only a minimal effect, because variation from \emph{crossover} was limited when the population had \emph{matured}, consisting of agent aggregations identical or very similar to one another. It should also be noted that the stability of the system is different to its performance, because although showing no instability with mutation rates below 60\% (inclusive), it only reached the \emph{maximum macro-state} $M_{max}$ with a mutation rate of 10\% or above, while at 0\% it was stable at a sub-optimal macro-state.

\section{Conclusions}

Our extension of the Chli-DeWilde definition of stability was developed to provide a greater understanding of \acp{MAS} with \emph{evolutionary dynamics}, specifically evolving agent populations, including our \emph{Digital Ecosystem}. We then built upon this to construct an entropy-based definition for the \emph{degree of instability} (entropy of the limit probabilities), which was also used to perform a \emph{stability analysis} of a simulated evolving agent population. Furthermore, our \emph{degree of instability} provides a definition for the \emph{level of stability}, applicable to \acp{MAS} with or without evolutionary dynamics.

Collectively, the experimental results confirm that Chli-Dewilde stability has been successfully extended to evolving agent populations, while our definition for the \emph{degree of instability} provides a macroscopic value to characterise the level of \emph{stability}. These findings also support the proposition that Chil-DeWilde stability can be widely applied to different types (classes) of \acp{MAS}, including our Digital Ecosystem, which is unique, and for which our simulations have shown some of its properties.

Our extended Chli-DeWilde stability is a useful \emph{tool} for analysing \acp{MAS}, with or without evolutionary dynamics, providing an effective understanding and quantification to help better understand such systems. Overall an insight has been achieved into the stability of \acp{MAS} with \emph{evolutionary dynamics}, including our Digital Ecosystem, which is a first step in being able to control such systems.

\section{Acknowledgments}

The authors would like to thank the following for encouragement and suggestions; Dr Paolo Dini of the London School of Economics and Political Science, Dr Maria Chli of Aston University, and Dr Jon Rowe of the University of Birmingham. This work was supported by the EU-funded \acs{OPAALS} Network of Excellence (NoE), Contract No. FP6/IST-034824.

\bibliographystyle{abbrv}
\bibliography{references} 

\begin{thebibliography}{10}

\bibitem{mspaperAbramov2001}
V.~Abramov, N.~Szirbik, J.~Goossenaerts, T.~Marwala, P.~De~Wilde, et~al.
\newblock Ontological basis for open distributed multi-agent system.
\newblock In {\em Symposium on Adaptive Agents and Multi-Agent Systems}, pages
  33--43. Society for the Study of Artificial Intelligence and Simulation of
  Behaviour, 2001.

\bibitem{mspaper9Balakrishnan1997}
H.~Balakrishnan, M.~Stemm, S.~Seshan, and R.~Katz.
\newblock Analyzing stability in wide-area network performance.
\newblock In S.~Leutenegger, editor, {\em International Conference on
  Measurement and Modeling of Computer Systems}, pages 2--12. ACM Press, 1997.

\bibitem{thesis}
G.~Briscoe.
\newblock {\em Digital Ecosystems}.
\newblock PhD thesis, Imperial College London, 2009.

\bibitem{javaOne}
G.~Briscoe, M.~Chli, and M.~Vidal.
\newblock {C}reating a {D}igital {E}cosystem: {S}ervice-oriented architectures
  with distributed evolutionary computing ({BOF}-0759).
\newblock In {\em JavaOne Conference}. Sun Microsystems, 2006.

\bibitem{bionetics}
G.~Briscoe and P.~{De Wilde}.
\newblock Digital {E}cosystems: Evolving service-oriented architectures.
\newblock In {\em Conference on Bio Inspired Models of Network, Information and
  Computing Systems}. IEEE Press, 2006.

\bibitem{dbebkpub}
G.~Briscoe and S.~Sadedin.
\newblock Digital {B}usiness {E}cosystems: Natural science paradigms.
\newblock In F.~Nachira, A.~Nicolai, P.~Dini, M.~Le~Louarn, and
  L.~Rivera~Le{\'o}n, editors, {\em Digital {B}usiness {E}cosystems}, pages
  48--55. European {C}ommission, 2007.

\bibitem{de07oz}
G.~Briscoe, S.~Sadedin, and G.~Paperin.
\newblock Biology of applied digital ecosystems.
\newblock In {\em Digital Ecosystems and Technologies Conference}, pages
  458--463. IEEE, 2007.

\bibitem{chlithesis}
M.~Chli.
\newblock {\em Convergence and Interactivity of Multi-Agent Systems}.
\newblock PhD thesis, Imperial College London, 2006.

\bibitem{chli2}
M.~Chli, P.~De~Wilde, et~al.
\newblock Stability of multi-agent systems.
\newblock In E.~Santos~Jr and P.~Willett, editors, {\em International
  Conference on Systems, Man, and Cybernetics}, pages 551--556. IEEE Press,
  2003.

\bibitem{msthesisCoxMiller1972}
D.~Cox and H.~Miller.
\newblock {\em The Theory of Stochastic Processes}.
\newblock CRC Press, 1977.

\bibitem{mspaperDeWilde1999a}
P.~De~Wilde, H.~Nwana, and L.~Lee.
\newblock Stability, fairness and scalability of multi-agent systems.
\newblock {\em International Journal of Knowledge-Based Intelligent Engineering
  Systems}, 3:84--91, 1999.

\bibitem{dini2008bid}
P.~Dini, G.~Lombardo, R.~Mansell, A.~Razavi, S.~Moschoyiannis, P.~Krause,
  A.~Nicolai, and L.~Rivera~Le{\'o}n.
\newblock Beyond interoperability to digital ecosystems: regional innovation
  and socio-economic development led by {SME}s.
\newblock {\em International Journal of Technological Learning, Innovation and
  Development}, 1:410--426, 2008.

\bibitem{eibenAarts}
A.~Eiben, E.~Aarts, and K.~Van~Hee.
\newblock Global convergence of genetic algorithms: A {M}arkov chain analysis.
\newblock In H.~Schwefel and R.~Manner, editors, {\em Parallel Problem Solving
  from Nature}, pages 4--12. Springer, 1991.

\bibitem{goldberg2}
D.~Goldberg and P.~Segrest.
\newblock Finite {M}arkov chain analysis of genetic algorithms.
\newblock In J.~Grefenstette, editor, {\em International Conference on Genetic
  Algorithms and their application}, pages 1--8. Lawrence Erlbaum Associates,
  1987.

\bibitem{mspaperLee1998}
L.~Lee, H.~Nwana, D.~Ndumu, and P.~De~Wilde.
\newblock The stability, scalability and performance of multi-agent systems.
\newblock {\em BT Technology Journal}, 16:94--103, 1998.

\bibitem{lin1994cgp}
S.~Lin, W.~Punch~III, and E.~Goodman.
\newblock Coarse-grain parallel genetic algorithms: categorization and new
  approach.
\newblock In {\em Symposium on Parallel and Distributed Processing}, pages
  28--37. IEEE Press, 1994.

\bibitem{mabu2007gbe}
S.~Mabu, K.~Hirasawa, and J.~Hu.
\newblock A graph-based evolutionary algorithm: Genetic network programming
  (gnp) and its extension using reinforcement learning.
\newblock {\em Evolutionary Computation}, 15:369--398, 2007.

\bibitem{ecpaper}
P.~Marrow.
\newblock Nature-inspired computing technology and applications.
\newblock {\em BT Technology Journal}, 18:13--23, 2000.

\bibitem{mspaperMarwala2001}
T.~Marwala, P.~De~Wilde, et~al.
\newblock Scalability and optimisation of a committee of agents using genetic
  algorithm.
\newblock In D.~Campbell and C.~Fyfe, editors, {\em International ICSC
  Symposium Soft Computing and Intelligent Systems For Industry}. ICSC-NAISO
  Academic Press, 2001.

\bibitem{moore1996}
J.~Moore.
\newblock {\em The Death of Competition: Leadership and Strategy in the Age of
  Business Ecosystems}.
\newblock Harvard Business School Press, 1996.

\bibitem{moreau2005sms}
L.~Moreau.
\newblock Stability of multiagent systems with time-dependent communication
  links.
\newblock {\em IEEE Transactions on Automatic Control}, 50:169--182, 2005.

\bibitem{nix}
A.~Nix and M.~Vose.
\newblock Modeling genetic algorithms with {M}arkov chains.
\newblock {\em Annals of Mathematics and Artificial Intelligence}, 5:79--88,
  1992.

\bibitem{msthesisNorris1997}
J.~Norris.
\newblock {\em {M}arkov Chains}.
\newblock Cambridge University Press, 1997.

\bibitem{masOverviewPaper}
H.~Nwana.
\newblock Software agents: An overview.
\newblock {\em Knowledge Engineering Review}, 11:205--244, 1996.

\bibitem{olfatisaber2007cac}
R.~Olfati-Saber, J.~Fax, and R.~Murray.
\newblock Consensus and cooperation in networked multi-agent systems.
\newblock {\em Proceedings of the IEEE}, 95:215--233, 2007.

\bibitem{moaspaper}
V.~Pham and A.~Karmouch.
\newblock Mobile software agents: an overview.
\newblock {\em IEEE Communications Magazine}, 36:26--37, 1998.

\bibitem{rudolph}
G.~Rudolph.
\newblock Convergence analysis of canonical genetic algorithms.
\newblock {\em IEEE Transactions on Neural Networks}, 5:96--101, 1994.

\bibitem{rudolph1998fmc}
G.~Rudolph.
\newblock Finite {M}arkov chain results in evolutionary computation: A tour
  d'horizon.
\newblock {\em Fundamenta Informaticae}, 35:67--89, 1998.

\bibitem{mspaperSimoesMarques2003}
M.~Simoes-Marques, P.~Mariano, R.~Ribeiro, L.~Correia, M.~Chli, P.~De~Wilde,
  V.~Abramov, and J.~Goosenaerts.
\newblock Contributions to adaptable agent societies.
\newblock In {\em Emerging Technologies and Factory Automation}, pages
  354--361. IEEE Press, 2003.

\bibitem{smith2000eec}
R.~Smith, C.~Bonacina, P.~Kearney, and W.~Merlat.
\newblock Embodiment of evolutionary computation in general agents.
\newblock {\em Evolutionary Computation}, 8:475--493, 2000.

\bibitem{smith1998fec}
R.~Smith and N.~Taylor.
\newblock A framework for evolutionary computation in agent-based systems.
\newblock In J.~Glasgow, editor, {\em International Conference on Intelligent
  Systems}, pages 221--224. AAAI Press, 1998.

\bibitem{mspaper5ThomasSycara1998}
J.~Thomas and K.~Sycara.
\newblock Heterogeneity, stability, and efficiency in distributed systems.
\newblock In Y.~Demazeau, editor, {\em International Conference on Multi Agent
  Systems}, pages 293 -- 300. IEEE Press, 1998.

\bibitem{swn1}
D.~Watts and S.~Strogatz.
\newblock Collective dynamics of `small-world' networks.
\newblock {\em Nature}, 393:440--442, 1998.

\bibitem{weiss1999msm}
G.~Weiss.
\newblock {\em Multiagent Systems: A Modern Approach to Distributed Artificial
  Intelligence}.
\newblock MIT Press, 1999.

\end{thebibliography}
\end{document}